\documentclass[runningheads]{llncs}

\usepackage[year=2026]{CVCONF}
\usepackage[width=122mm,left=12mm,paperwidth=146mm,height=193mm,top=12mm,paperheight=217mm]{geometry}

\usepackage{setspace}
\usepackage{etoolbox} 
\AtBeginDocument{\setstretch{0.95}\fontsize{9pt}{10.8pt}\selectfont}
\AtBeginEnvironment{table}{\setstretch{0.95}}

\usepackage{libertine}
\usepackage[libertine]{newtxmath}
\usepackage[scaled=0.9]{inconsolata}

\usepackage{CVCONFabbrv}
\usepackage{graphicx}
\usepackage{booktabs}
\usepackage{multirow}
\usepackage{capt-of}
\usepackage{float}
\usepackage[accsupp]{axessibility}  

\usepackage{algorithm}
\usepackage{algpseudocode}

\usepackage[breaklinks,colorlinks,citecolor=CVCONFblue]{hyperref}

\usepackage{orcidlink}

\usepackage{xcolor}

\begin{document}

\title{UniBioTransfer: A Unified Framework for Multiple Biometrics Transfer}

\titlerunning{ }

\author{
  Caiyi Sun\inst{1}\textsuperscript{*}\and
  Yujing Sun\inst{2}\textsuperscript{*}\and
  Xiangyu Li\inst{2}\and
  Yuhang Zheng\inst{2}\and
  Yiming Ren\inst{2,3}\and
  Jiamin Wang\inst{3}\and
  Yuexin Ma\inst{3}\and
  Siu-Ming Yiu\inst{1}\textsuperscript{$\dagger$}
}
\authorrunning{ }
\institute{
  The University of Hong Kong\and
  Digital Trust Centre, Nanyang Technological University\and
  ShanghaiTech University
}

\makeatletter
\let\@oldmaketitle\@maketitle
\renewcommand{\@maketitle}{
  \@oldmaketitle
  \begin{center}
    \includegraphics[width=0.93\linewidth]{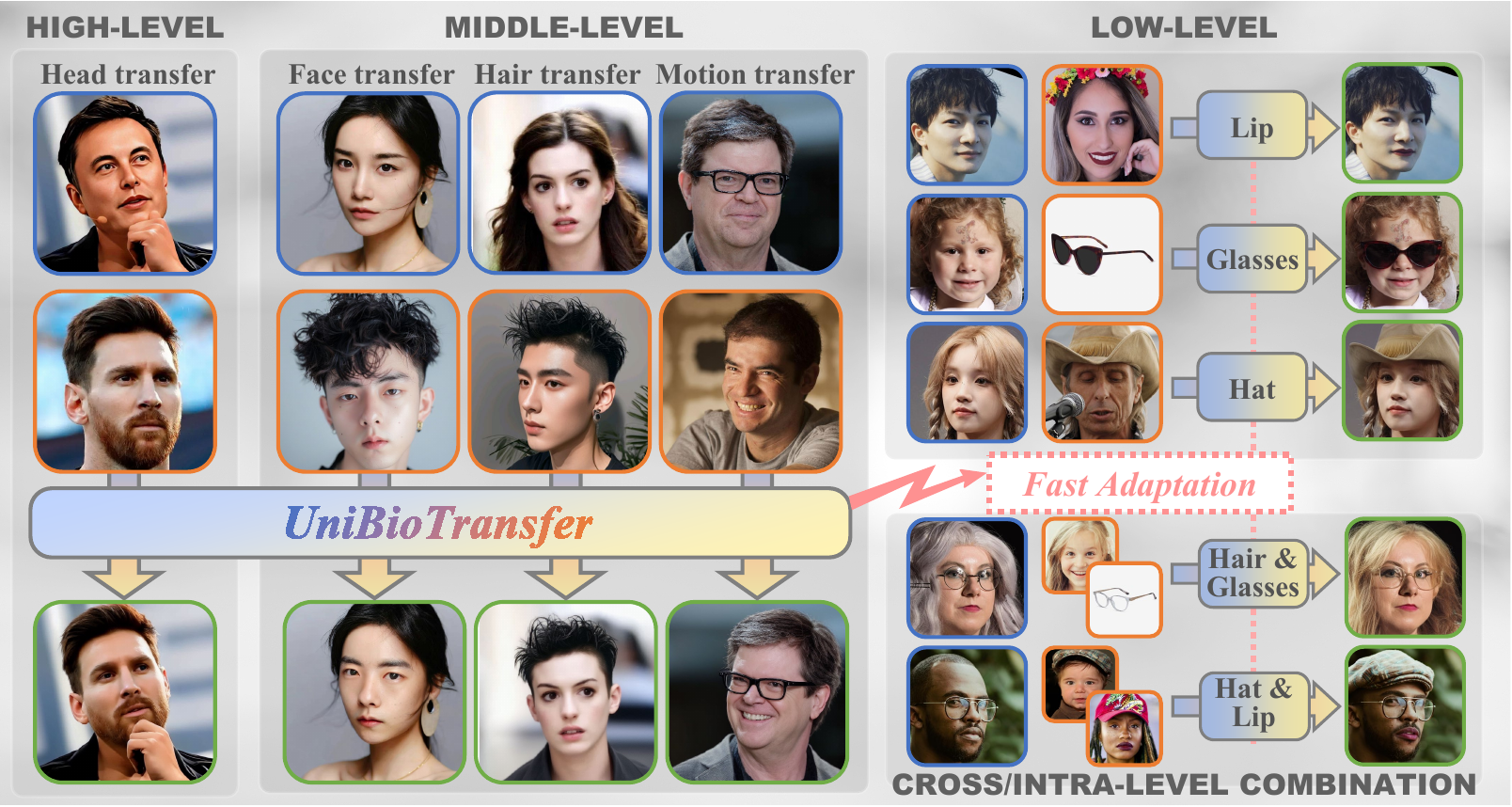}
  \end{center}
  \vspace{-1.0ex}
  \refstepcounter{figure}\normalfont Figure~\thefigure.
  \textbf{UniBioTransfer} is the first unified framework capable of handling all four complicated and typical high-level and mid-level deepface generation tasks within the same model, while also generalizing efficiently to novel low-level and cross-level/intra-level compositional transfer tasks by conducting minimal fine-tuning with $<20\%$ data and $<10\%$ training cost.
  (\textcolor[RGB]{68, 114, 196}{blue border}: target image, \textcolor[RGB]{237, 125, 49}{orange border}: reference images, \textcolor[RGB]{113, 173, 72}{green borders}: our transferred results).
  \label{fig:teaser}
  \par\vspace{-1.0ex}
}
\makeatother

\maketitle
\begingroup
\makeatletter
\def\@fnsymbol#1{\ifcase#1\or *\or \ensuremath{\dagger}\else\@ctrerr\fi}
\makeatother
\renewcommand{\thefootnote}{\fnsymbol{footnote}}
\setcounter{footnote}{0}
\footnotetext[1]{Equal contribution.}
\footnotetext[2]{Corresponding author.}
\endgroup

\begin{abstract}

Deepface generation has traditionally followed a task-driven paradigm, where distinct tasks (e.g., face transfer and hair transfer) are addressed by task-specific models.
Nevertheless, this single-task setting severely limits model generalization and scalability. 
A unified model capable of solving multiple deepface generation tasks in a single pass represents a promising and practical direction, yet remains challenging due to data scarcity and cross-task conflicts arising from heterogeneous attribute transformations.
To this end, we propose \textbf{UniBioTransfer}, the first unified framework capable of handling both conventional deepface tasks (e.g., face transfer and face reenactment) and shape-varying transformations (e.g., hair transfer and head transfer).
Besides, \textit{UniBioTransfer} naturally generalizes to unseen tasks, like lip, eye, and glasses transfer, with minimal fine-tuning.
Generally, \textit{UniBioTransfer} addresses data insufficiency in multi-task generation through a unified data construction strategy, including a swapping-based corruption mechanism designed for spatially dynamic attributes like hair. It further mitigates cross-task interference via an innovative BioMoE, a mixture-of-experts based model coupled with a novel two-stage training strategy that effectively disentangles task-specific knowledge.
Extensive experiments demonstrate the effectiveness, generalization, and scalability of \textit{UniBioTransfer}, outperforming both existing unified models and task-specific methods across a wide range of deepface generation tasks. 
Project page is at \url{https://scy639.github.io/UniBioTransfer.github.io/}
\end{abstract}

\vspace{-1.5em}
\section{Introduction}
\label{sec:intro}
\vspace{-0.5em}

DeepFace generation centers on the transfer (also referred to as \textbf{swapping}) of biometric attributes within the head region from a reference image to the target image. This spans from high-level whole-head to mid-level biometrics like the face and hair, and low-level fine-grained details like facial parts (e.g., eyes and lips), and accessories.
The current landscape of DeepFace research remains largely task-specific, focusing on individual applications such as head transfer~\cite{fewshot_headswap_cvpr22,ghost20}, 
face transfer~\cite{diffswap_cvpr23,idconstrained_fswapping}, 
motion transfer (also known as face reenactment)~\cite{liveportrait,echomimic_aaai25}, 
and hair transfer~\cite{stable_hair_aaai25,hairfusion_aaai25}. 
Multi-task frameworks~\cite{reface_wacv25} also require extensive retraining for each new task. In general, such approaches exhibit poor generalization, as models optimized for a single task often struggle to transfer knowledge across related tasks, and lack practicality for real-world deployment, especially when scaling to diverse deepface applications.

These limitations highlight the need for a unified framework to efficiently handle multiple generation tasks within a single model. We observe that most deepface generation tasks share some correlations, as they fundamentally involve head region feature transfer. This observation motivates the design of a framework that can leverage inter-task knowledge to jointly address and benefit multiple tasks. Such an approach improves practicality and scalability by eliminating the need to train and maintain separate task-specific models.

\begin{figure}[t]
  \begin{center}
    \includegraphics[width=0.8\textwidth]{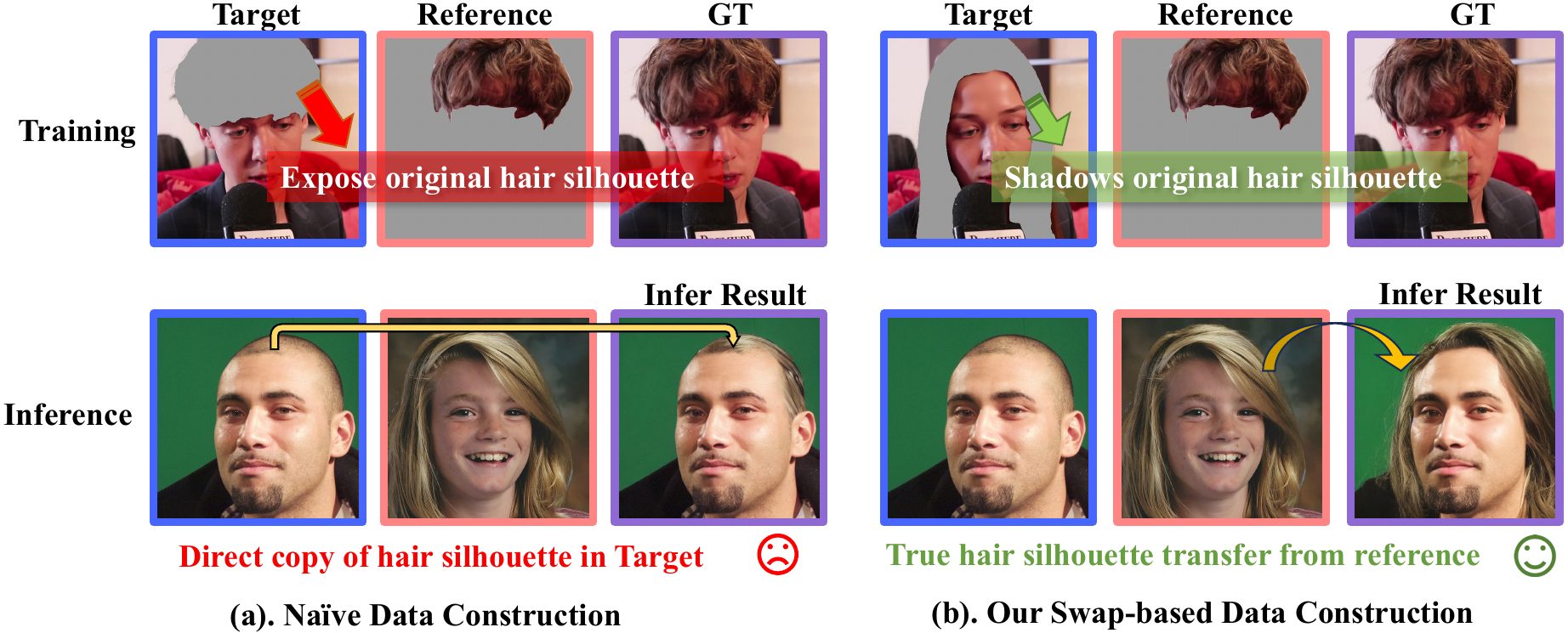}
  \end{center}
   \vspace{-1em}
  \caption{
Limitations of traditional mask-based strategy for attributes with significant structural changes (e.g., hair transfer). 
Masking exposes ground-truth geometry (a-top), causing models to learn only inpainting rather than true shape transfer. Our swapping-based strategy removes silhouette information in the target (b-top), forcing the network to transfer shape from the reference.
}
  \label{fig:mask_issue}
  \vspace{-1.5em}
\end{figure}

Nevertheless, designing a robust unified model capable of handling multiple complex deepface tasks remains highly challenging.
\textbf{\textit{Data Dilemma.}}
The scarcity of paired data for deepface generation tasks necessitates a self reconstruction training paradigm, where the target, reference, and ground-truth images are derived from the same identity. Typically, a target is constructed by masking out specific regions of the ground truth, a strategy proven effective in previous works\cite{reface_wacv25}. 
However, this naive mask-based strategy fails for large structural changes because mask silhouette leaks geometry, causing models to simply inpaint within the mask rather than perform true shape transfer (Fig.~\ref{fig:mask_issue}). Alternative strategies, such as StableHair~\cite{stable_hair_aaai25} with its error-prone hair-removal model (Fig.~\ref{fig:comp}c), HS-Diffusion~\cite{hs_diffusion} requiring extra networks compromising unified framework design, and HairFusion~\cite{hairfusion_aaai25} relying on aggressive masking and artifact-prone blending—each introduce substantial limitations (Fig.~\ref{fig:comp}c).
\textbf{\textit{Cross-Task Conflict Dilemma.}}
Meanwhile, although deepface tasks all focus on manipulation of the human head region,
they also exhibit distinct task-specific objectives and feature distributions. For instance, face transfer emphasizes identity preservation, while face reenactment focuses on pose\&expression transfer, and hair or head transfer requires structural adaptation. 
These divergent objectives create inherent conflicts when jointly optimized within a single model.
Naïvely training a shared network across multiple tasks often leads to gradient interference, where updates beneficial for one task degrade the performance of others.

Hence, we propose \textbf{UniBioTransfer}, a unified deepface generation framework that introduces innovations at both the data and model levels to address the aforementioned challenges.
To bridge the data gap, we propose a unified data construction strategy. We first define all tasks as transferring a set of attributes (e.g., hair, identity) from a reference image $I_\text{ref}$ to a target image $I_\text{tgt}$ while preserving the remaining attributes of the target (see Section~\ref{sec:method}).
Based on this, we formulate a unified data construction paradigm termed \textbf{attribute corruption}: starting from a ground-truth image $I_\text{gt}$, we divide its attributes into those to be transferred and those to be preserved, then corrupt each set to create $I_\text{tgt}$ and $I_\text{ref}$.
For relatively-static attributes like facial identity, simple masking suffices to effectively corrupt the identity information in the ground-truth image.
However, for spatially-dynamic attributes where masks often leak geometry, we propose a novel \textbf{Swapping-based Corruption} strategy to prevent trivial solutions in tasks with large shape variations.
Specifically, we leverage an off-the-shelf generative model to transfer these regions with novel geometries, compelling the network to learn genuine shape transfer from the reference (Fig.~\ref{fig:mask_issue}b) rather than falling into a trivial inpainting-like solution (Fig.~\ref{fig:mask_issue}a).

For the cross-task conflict dilemma, we propose \textbf{BioMoE}, an MoE-based architecture tailored for our task, and a \textbf{Two-Stage Training Strategy}. Unlike a conventional "black-box" router that relies solely on input tokens, our \textbf{BioMoE} additionally incorporates spatial structural cues by providing the router with each token's relative position to a set of facial landmarks, enabling the routed experts to specialize in distinct anatomical regions (e.g., identity-rich centers vs. geometry-heavy boundaries). To further mitigate gradient interference in multi-task learning and to facilitate adaptive capacity allocation across tasks, we devise a \textbf{two-stage training strategy} for \textbf{BioMoE}: Stage I (Task-Specific Pre-training) aims to pre-train individual task-specific modules with gradients isolated to rapidly 
obtain a superior starting point that prevents early gradient clashes. In Stage II (Task-Unified Fine-tuning), the model is jointly optimized, transitioning from isolated task weights to our final BioMoE-based network.


We summarize our contribution as follows:

     \noindent \textbf{1).} \quad We present the first framework, \textbf{UniBioTransfer},  to effectively unify complex tasks with shape-varying transformations (e.g., head and hair transfer), within a single architecture.
    
    
     \noindent \textbf{2).}\quad We propose a \textbf{unified data construction strategy} to solve the data insufficiency issue in multi-generation tasks,  and a novel \textbf{BioMoE} with \textbf{Two-Stage Training Strategy}, tailored for our tasks with reduced inter-task interference and improved cross-task collaborative generation.
    
    \noindent \textbf{3).}\quad We achieve state-of-the-art performance across a wide range of representative deepface generation tasks, 
    showing strong generalization and scalability.

\vspace{-0.5em}
\section{Related Works}
\label{sec:related}

\vspace{-0.2em}
\subsection{DeepFace Generation}

\noindent\textbf{Task-Specialized Methods}
Early task-specialized methods largely relied on GANs and warping, offering efficiency but struggling with fidelity.
Recent diffusion approaches substantially improve identity preservation, texture realism, and compositional control across tasks such as face transfer \cite{e4s,diffswap_cvpr23,hifivfs,idconstrained_fswapping,reface_wacv25,selfswapper_eccv24,dreamID,canonswap_iccv25,dynamicface_iccv25,3daware_face_swapping_cvpr23,fuseanypart_neurips24}, motion transfer (face reenactment) \cite{liveportrait,magicportrait,realportrait_aaai25,hunyuanportrait_cvpr25,emojidiff,follow_your_emoji,echomimic_aaai25,megactor_sigma,skyreels_a1,megactor_sigma_aaai25}, hair transfer \cite{hairclip_cvpr22,hairclipv2_iccv23,hairfastgan_neurips24,hairfusion_aaai25,stable_hair_aaai25}, and head transfer \cite{fewshot_headswap_cvpr22,ghost20,hs_diffusion,zeroshot_headswap_cvpr25}.
Overall, such methods suffer from limited generalization—models trained for one task transfer poorly to others. This confines their applicability to single-task scenarios, which limits their versatility as the range of deepface applications expands.

\noindent\textbf{Multi-Task Methods}
Recognizing the limitations of task-specific models, several works have aimed to unify different DeepFace tasks. Early efforts~\cite{uniface__unified_reenact_swap_eccv22,faceadapter_eccv24,unifacepp_ijcv25,towards_consistent_face_editing_arxiv25}, such as FaceAdapter~\cite{faceadapter_eccv24} and RigFace~\cite{towards_consistent_face_editing_arxiv25}, unified face swapping and face reenactment by exploiting their shared underlying pattern: the recombination of identity and motion (pose and expression) from target and reference faces.
Nevertheless, their heavy dependence on 3DMMs restricts their applicability to tasks involving significant shape deformations beyond facial regions, a constraint that our approach effectively addresses.

\vspace{-0.5em}
\subsection{Reference-Based Image Editing}
\vspace{-0.3em}
Beyond the DeepFace domain, general image editing models have made significant strides.
Instruction-driven editors offer broad, promptable control without task-specific tuning, but the visual quality can be inconsistent. Even with stronger preservation in recent large vision-language models like Kontext~\cite{flux_kontext}, Qwen-Image-Edit~\cite{qwen_image_edit}, reference-based image editing\cite{instructpix2pix,pix2pix_zero,anyedit_cvpr25,masactrl_iccv23,prompt2prompt_iclr23,text2videozero_iccv23} usually fails to 
precisely transfer the reference attribute to the target~\cite{xia2025dreamomni2}.
Consequently, these general editing methods are rarely utilized in practical deepface generation because they are unreliable under multi-image (reference-based) inputs (see Sec. B.1 in supplementary material). However, they can be effectively leveraged for our training data construction, as detailed in Sec.~\ref{sec:swapping_based_corruption}.

\vspace{-0.5em}
\subsection{Mixture-of-Experts (MoE) and Multi-task Learning}
\vspace{-0.3em}

Mixture-of-Experts (MoE)  is widely used across a wide range of fields (e.g., recommendation system~\cite{mmoe,ple}, and Large Language Model~\cite{switch_transformer,gshard,deepseek_moe,expert_choice}).

In vision field, recent works~\cite{uni_controlnet,sun2024anycontrol,t2i_adapter} introduce MoE to handle multi-control image generation, but they are not inherently for multi-task learning as they focus on the fusion of conditions which are complementary and additive, rather than resolving fundamental cross-task conflicts.
Many works~\cite{adamv_moe,uni_perceiver_moe,transforming_vit_neurips24,unihcp_cvpr23,unimed_neurips24,faceptor_eccv24} successfully introduce MoE to address cross-task challenges using dynamic routing.
Despite their success, these models are not directly suitable for our objectives. Adapting "all-routed" MoE architectures like AdaMV-MoE~\cite{adamv_moe} to our backbone is impractical: since state-of-the-art generation models are predominantly dense, retraining a sparse MoE from scratch requires prohibitive resources, and upcycling often incurs significant performance loss.
M3DT~\cite{mastering_mtrl} employs full-scale experts identical to the backbone FFN for each task group, yet this would lead to a prohibitive memory requirement in our case, as our backbone model~\cite{stableDiffusion} is already massive.
We thus propose a custom Mixture-of-Experts (BioMoE) that maintains compatibility with pre-trained dense models like Stable Diffusion~\cite{stableDiffusion} while enabling precise specialization and parameter efficiency.

\vspace{-0.5em}
\section{Method}
\label{sec:method}
\vspace{-0.5em}

A unified model capable of handling multiple deepface transfer tasks is promising yet highly challenging. 
On one hand, at the data level, constructing effective training pairs for shape-varying attributes (e.g., hair, glasses, hats) is difficult. Conventional masking or augmentation-based strategies often cause shape information leakage, hindering the model from learning genuine geometric transformations. 
On the other hand, at the model level, naively sharing all parameters across heterogeneous tasks leads to severe task interference.

To address these challenges, we introduce \textbf{UniBioTransfer}, a unified framework for multi-level deepface generation tasks. To be specific, (1) we propose a novel \textbf{Unified Data Construction Strategy} tailored for UniBioTransfer, which effectively alleviates the data insufficiency issue in multi-level deepface generation tasks. In addition, (2) we present a custom MoE-based network \textbf{BioMoE} that aims to reduce inter-task interference. Finally, (3) we design a \textbf{Two-Stage Training Strategy} to leverage task-specific pre-training and task-unified fine-tuning. This strategy enables our BioMoE to maintain outstanding performance across multiple generation tasks simultaneously.

\textit{Problem Definition:} \quad
We formulate various deepface tasks as swapping a set of attributes $X$ (e.g., face identity, hair, pose, expression, skin tone) from a reference image $I_\text{ref}$ onto a target image $I_\text{tgt}$, while preserving the remaining attributes $Y$ from $I_\text{tgt}$. The desired output is $I_\text{out} = X_\text{ref} \cup Y_\text{tgt}$.

\textit{Pipeline Overview:} \quad
We adopt Stable Diffusion v1.5\cite{stableDiffusion} as the backbone.
As shown in Fig.~\ref{fig:model}, given a target image $I_\text{tgt}$ and a reference image $I_\text{ref}$, we first mask out irrelevant attributes in both.  
We extract semantic features of masked $I_\text{ref}$ via the CLIP encoder~\cite{openai_clip} and identity features (for tasks where $I_\text{ref}$ provides identity attributes like face/head transfer) via the ArcFace model~\cite{deng2019arcface}. 
An MLP then projects these raw features into tokens for cross-attention conditioning in UNets. 
We form blended landmarks to guide motion, taking pose\&expression from $I_\text{ref}$ (for motion transfer) or $I_\text{tgt}$ (for other tasks) and identity from the counterpart.  
The masked $I_\text{tgt}$  is encoded to a latent and concatenated with noise along the channel dimension as input to the main UNet.  
The masked $I_\text{ref}$  flows through the refNet (Reference-UNet), and its tokens are injected via cross-attention into the main UNet.
We use only the first half of the refNet to improve efficiency.
After iterative denoising, the latent is decoded back into pixel space by VAE decoder to produce the final result $I_\text{out}$.
Refer to Sec. A.1 in supplementary material for details.

\vspace{-0.5em}
\subsection{A Unified Data Construction Strategy}
\label{sec:method_data}
\vspace{-0.5em}

The lack of paired data in deepface generation necessitates self-reconstruction training, meaning the $I_\text{tgt}$, $I_\text{ref}$, and $I_\text{gt}$ come from the same identity. However, this setup risks trivial learning for shape-varying tasks, such as hair or head transfer, with models preserving spatial attributes in $I_\text{tgt}$ rather than transferring them from $I_\text{ref}$ (Fig.~\ref{fig:mask_issue}). Task-specific solutions for shape-varying tasks remain insufficient: StableHair~\cite{stable_hair_aaai25} relies on a bald converter, introducing error accumulation; HS-Diffusion~\cite{hs_diffusion} adds a mask predictor, breaking model unification; and HairFusion~\cite{hairfusion_aaai25} uses overly aggressive masking, causing background distortion and boundary artifacts (Fig.~\ref{fig:comp}b). 

\begin{figure}[t]
  \begin{center}
    \includegraphics[width=0.6\textwidth]{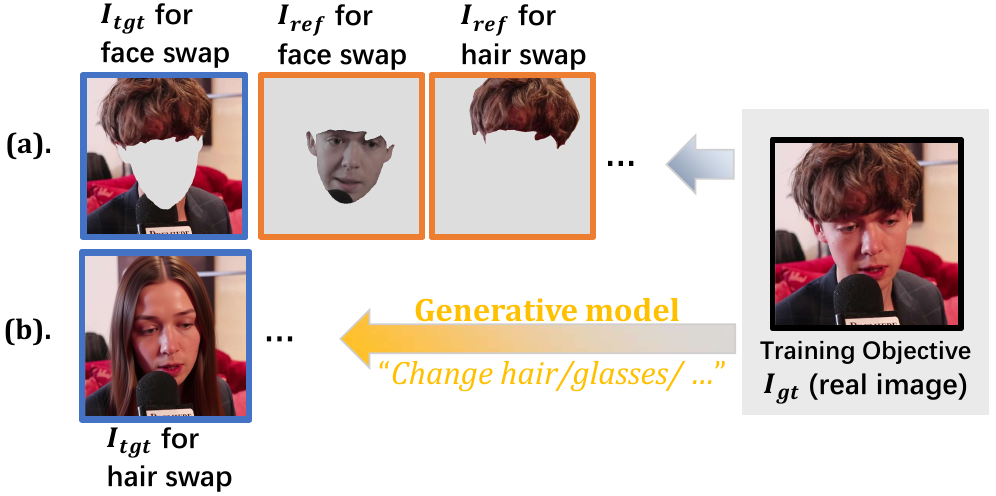}
  \end{center}
  \vspace{-1em}
  \caption{
Our unified data corruption strategy for different attribute types. (a) Relative-static attributes: the target is constructed by simple masking or data augmentation of the GT image. (b) Spatially-dynamic attributes: we utilize our swapping-based corruption strategy, which employs an off-the-shelf generative model to replace specific attributes in the GT with arbitrary novel variations, preventing shape leakage from mask boundaries.
\vspace{-1em}
}\label{fig:general_data_construction}
\end{figure}

\begin{figure*}[t]
  \centering
  \begin{center}
    \includegraphics[width=1\textwidth]{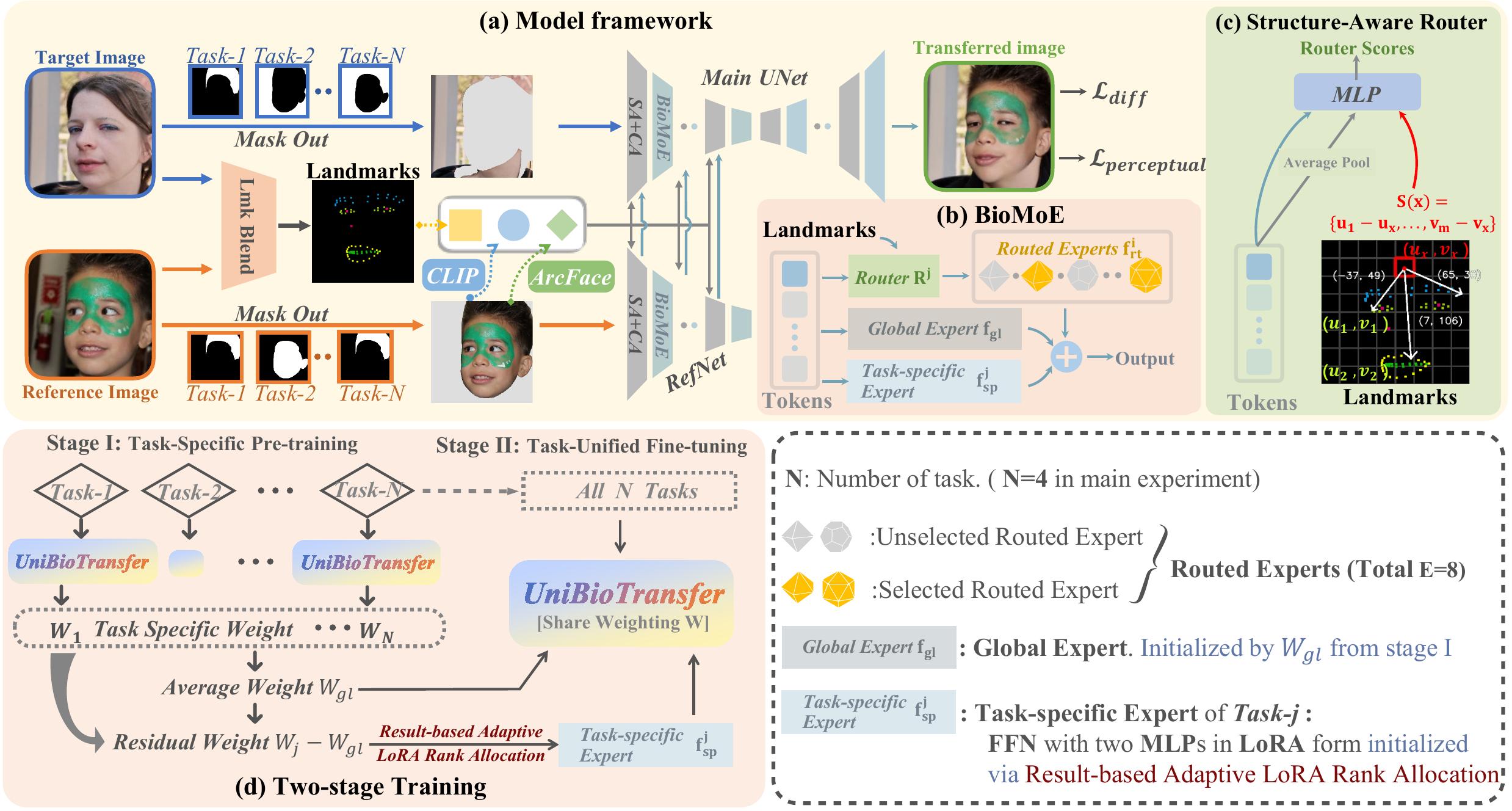}
  \end{center}
\vspace{-1.5em}
  \caption{\textbf{UniBioTransfer} architecture overview. 
(a) Overall framework. 
(b) We introduce an MoE-enhanced Feed Forward Network (FFN). 
(c) Expert selection is guided by a Structure-Aware Router. 
(d) The entire system is optimized using a two-stage training strategy designed to stabilize routing and promote expert specialization.
}\label{fig:model}
\vspace{-1.5em}
\end{figure*}

To address these limitations, we introduce our unified data construction strategy (Fig.~\ref{fig:general_data_construction}). 
We formulate the data construction process for various deepface generation tasks as \textbf{attribute corruption}: beginning with a ground-truth image $I_\text{gt} = X_\text{gt} \cup Y_\text{gt}$, where $X_\text{gt}$ and $Y_\text{gt}$ are the ground-truth attributes corresponding to the sets $X$ and $Y$, the goal of data construction is to synthesize the training pair $(I'_\text{tgt}, I'_\text{ref})$ by strategically corrupting attributes in $I_\text{gt}$. Specifically, the target image $I'_\text{tgt}$ is constructed by corrupting the attributes $X_\text{gt}$ to $X_\text{corrupted}$ while preserving $Y_{gt}$, resulting in $I'_\text{tgt} = X_\text{corrupted} \cup Y_\text{gt}$. Conversely, the reference image $I'_\text{ref}$ is formed by corrupting $Y_\text{gt}$ to $Y_\text{corrupted}$ while preserving $X_\text{gt}$, yielding $I'_\text{ref} = X_\text{gt} \cup Y_\text{corrupted}$.
As a result, the constructed data pair $(I'_\text{tgt}, I'_\text{ref}) \rightarrow I_\text{gt}$ compels the model to learn the intended transfer behavior: to extract attribute set $X$ from $I'_\text{ref}$ and merge it with attribute set $Y$ from $I'_{tgt}$.

The critical challenge in this framework is ensuring that the corrupted attributes $X_{\text{corrupted}}$ do not leak the original geometry of $X_{gt}$. We classify attributes into relatively-static and spatially-dynamic, and discuss specific corruption methods for each:

\paragraph{\textbf{Relative-Static Attributes Corruption}} (Fig.~\ref{fig:general_data_construction}a).  \quad
Relative-static attributes include both structural and non-structural elements whose spatial locations and overall shapes remain largely consistent during transfer.
For structural attributes (e.g., face, eyes, nose), which have well-defined geometric boundaries, we apply mask-based corruption, a widely used and effective strategy that removes texture and color information, forcing the model to reconstruct these details from the reference.
For non-structural attributes (e.g., skin tone), we employ standard data augmentation techniques to introduce variation while preserving structure.

\paragraph{\textbf{Spatially-Dynamic Attributes Corruption}} (Fig.~\ref{fig:general_data_construction}b). \quad
Spatially-dynamic attributes include those where the shape is a critical component to be transferred, such as hair or accessories (e.g., hats, glasses). 
Simple masking is troublesome here, as the mask boundary itself leaks the ground-truth silhouette, allowing the model to find a trivial inpainting-like solution instead of learning shape transfer~\cite{reface_wacv25}. 
To address this, we propose a novel Swapping-Based Corruption Strategy. 

\textit{Swapping-Based Corruption Strategy.} \label{sec:swapping_based_corruption}
Specifically, we leverage an off-the-shelf generative model to replace the attribute in $I_\text{gt}$ with one from an arbitrary image or text prompt, thereby creating the corrupted target $I'_\text{tgt}$ as shown in Fig.~\ref{fig:general_data_construction}~\cite{qwen_image_edit}. 
This generates a target image with a completely different hair shape and color, ensuring the model cannot rely on the target's original hair shape and must learn to extract it entirely from the reference (Fig.~\ref{fig:mask_issue}). 
For novel spatially-dynamic attributes where no specialized model exists, a general-purpose image editing model will work.

\paragraph{Remarks.} 
The proposed data construction strategy can be seen as a way to transform imperfect generative models, those with limited preservation or transfer capabilities, into stronger models. The only requirement for the external generative model used in our swapping-based corruption is the ability to introduce diverse attribute variations. It does not need to transfer attributes from a reference image, as our corruption only modifies the target image, nor does it need to preserve non-target regions, 
since low-quality pairs are filtered out using preservation metrics, including background SSIM, identity similarity, pose and expression distance 
(see Sec. B.3 in Suppl. for examples of discarded and retained pairs).
Thus, even for tasks involving significant shape variations, our method enables the final model to achieve a preservation and transfer fidelity that surpasses the initial model used for data generation.
We discuss the self-evolution potential and provide an additional quantitative experiment in Sec. B.1(b) in Suppl.

\vspace{-0.5em}
\subsection{BioMoE}
\label{sec:method_model}
\vspace{-0.5em}
A unified model capable of handling multiple deepface tasks is highly desirable for efficiency but faces a key challenge: task interference due to disparate task requirements. For example, face transfer necessitates high-precision identity transfer within the central facial region, whereas hair/head transfers focus on geometric deformations at the peripheral boundaries. 
An alternative is to directly adopt the standard Mixture of Experts (MoE) architecture in LLM~\cite{deepseek_moe} and multi-task vision~\cite{adamv_moe} fields. 
However, this typically requires a pre-trained MoE image generation base model, whereas state-of-the-art image generation models are predominantly dense. 
Other approaches~\cite{mastering_mtrl} copy full-scale experts from the backbone FFN for each task group. Yet, this would lead to a prohibitive memory requirement, as the base model~\cite{stableDiffusion} for our tasks is already heavy.

To resolve these conflicting constraints, we introduce our \textbf{BioMoE}. We achieve an effective design with acceptable overhead by retaining only one identical FFN from the pretrained dense model~\cite{stableDiffusion} for knowledge sharing, while delegating dynamic and specialized knowledge to lightweight routed experts and task-specific experts.
Our proposed BioMoE module consists of three types of experts, balancing shared knowledge, task-specific knowledge, and dynamic specialization:

\textbf{1). Global Expert.} 
The original FFN structure from the pretrained model is preserved to serve as a common knowledge base that processes all input tokens, capturing the foundational paradigm of all tasks: extracting reference attributes and blending them into the target image.

\textbf{2). Routed Experts.} 
A pool of $E=8$ lightweight experts is shared across all tasks for dynamic and structure-aware specialization. For each input token, a task-specific router selects the top-$K$ experts from this pool to process the token. Each expert follows the FFN's two-layer MLP structure, but with a reduced inner dimension (e.g.,  reduced by a factor of $F=16$ compared to original FFN's dimension) for efficiency.

\textbf{3). Task-specific Experts.} 
To encode highly specialized information unique to each task's distinct requirements (e.g., precise identity process for face/motion transfer or geometric deformation for hair/head transfer), we append a task-specific expert for each task, efficiently implemented using Low-Rank Adaptation (LoRA).
Formally, the forward  process of BioMoE is as follows:
\vspace{-0.7em}
\begin{equation}
  f(x)=f_{\mathrm{gl}}(x) + \sum_{i=1}^{E}\mathcal{G}^{(j)}(x) \cdot   
  f_{\mathrm{rt}}^{(i)}(x)  + f_{\mathrm{sp}}^{(j)}(x),
  \label{eq:moe}
\end{equation}
where $f_{\mathrm{gl}}$ is the global shared FFN, $\mathcal{R}(x) \in \mathbb{R}^E$ contains the routing weights (with only top-$K$ non-zero values), $f_{\mathrm{rt}}^{(i)}$ denotes the $i$-th dynamically routed expert, and $f_{\mathrm{sp}}^{(j)}$ represents the task-specific expert of task-$j$. Meanwhile, 
\begin{equation}
\mathcal{G}^{(j)}(x)=\mathrm{top}_{K}\left(\mathrm{softmax}\left(  R^{(j)}(\operatorname{concat}(x,\operatorname{pool}(X),\mathcal{S}(x)))+\epsilon\right)\right),
\label{eq:router}
\end{equation}
in which the $\mathrm{top}_{K}$  operator sets all values to be zero except the largest $K$ values, $\mathcal{G} \in \mathbb{R}^E$ denotes the routing weights, $R^{(j)}$ is the task-specific router for task $j$, $\operatorname{concat}(x,\operatorname{pool}(X),\mathcal{S}(x))$ concatenates the token feature $x$, the pooled sequence features, and structure information $\mathcal{S}(x)$ of the token, $E$ is the number of experts, and $j$ is the task id.
$\epsilon$ is a noise term added during training to encourage exploration and improve expert load balancing.

\paragraph{\textbf{Structure-Aware Routing}} \quad
\label{sec:method_gate}
Different deepface generation tasks emphasize distinct semantic regions—for instance, face transfer and reenactment focus on facial identity, while hair transfer targets the scalp area. To exploit this spatial locality, we introduce a structure-aware router.
Unlike conventional MoE routers that rely solely on token features, our router is an MLP that also incorporates structure information (relative positions to each facial landmarks). This design enables the routing mechanism to dynamically select experts based on spatial context. For example, tokens corresponding to facial regions can be routed toward experts specialized in identity preservation, while tokens from the scalp region activate experts focusing on hair transfer.
The structure information $\mathcal{S}(x)$ for each token is computed as:
\begin{equation}
  \mathcal{S}(x) = \operatorname{concat}\big( u_1 - u_x, v_1 - v_x, \dots, u_M - u_x, v_M - v_x \big)
\end{equation}
where $(u_m, v_m)$ denotes the coordinates of the $m$-th landmark, $(u_x, v_x)$ is the spatial position of token $x$ in the feature map, and $M$ is the number of key landmarks.

\subsection{Two-Stage Training Strategy}\quad
\label{sec:model_2_stage_training}
A straightforward approach to training our model would be to initialize all components and jointly train on all tasks from the outset. 
However, this is suboptimal as: (1) the global expert contains the majority of the parameters, thus joint training from the start would induce significant gradient conflicts within this large shared module, slowing convergence; 
(2) the optimal rank for the task-specific experts can vary significantly depending on the task and layer complexity. 
Thus, we propose an efficient two-stage training strategy, which includes Task-Specific Pre-training and Task-Unified Fine-tuning to achieve more stable and specialized convergence.

\paragraph{\textbf{Stage I: Task-Specific Pre-training}}\quad
We first train a separate version of the model for each individual task using the same model structure. Specifically, a standard FFN is trained for each task separately, with gradients isolated between them and without routed experts. 
This stage serves a dual purpose: 1). allowing task-specific modules to rapidly achieve relatively optimal performance on each task, providing a high-quality initialization for stage II, and 2). establishing a comprehensive weight profile for each task capability, enabling our \textit{Result-Based Adaptive LoRA Rank Allocation}.

\textbf{Result-Based Adaptive LoRA Rank Allocation.} \quad
\label{sec:ada_lora}
A key mechanism in our two-stage training is the Result-Based Adaptive LoRA Rank Allocation,
which utilizes the weight from 1$_{st}$ stage to allocate parameter budgets in a dynamic and data-driven manner, ensuring each task receives its optimal capacity before the final joint optimization.
Unlike conventional "gradient-oriented" adaptive LoRA methods\cite{gradBased_ada_lora}, which are inherently "greedy" as they depend on short-term gradient information that can be noisy and misleading, we propose to utilize the pretrained weights from stage I, which reflect long-term task requirements. 
Specifically, we perform Singular Value Decomposition (SVD) on the residual matrix ($W_j - W_{\text{gl}}$) for each task-specific expert. We then select the rank $r$ such that the top $r$ singular values capture a predefined percentage (e.g., 20\%) of the total matrix energy. This adaptive approach automatically allocates more parameters (a higher rank) to tasks and layers that require more complex, specialized adaptation, while using a lower rank for simpler ones, resulting in a highly efficient and optimized architecture.

\paragraph{\textbf{Stage II: Task-Unified Fine-tuning}}\quad
We then construct our final MoE-based unified model.
In detail, we initialize the weights of the global expert in each MoE layer by averaging the weights of the corresponding FFNs from the model trained in stage one ($W_{\text{gl}} = \frac{1}{N}\sum_{j=1}^{N} W_j$). Next, the task-specific experts need to be initialized to capture the residual information for each task. For each task $j$, the task-specific expert's weights are initialized by fitting the difference between the weights of the task-specific FFN ($W_j$) and the averaged global expert ($W_{\text{gl}}$), where the LoRA rank is determined by the proposed Adaptive LoRA Rank Allocation in Sec.~\ref{sec:ada_lora}.

\paragraph{Loss Function.} \quad
Our training loss consists of a standard DDPM loss and image-space perceptual losses, please refer to Sec. A.2 in Suppl. for details.
 

\section{Experiments and Discussions}
\label{sec:experiments}

\begin{figure*}[t]
  \begin{center}
    \includegraphics[width=0.99\textwidth]{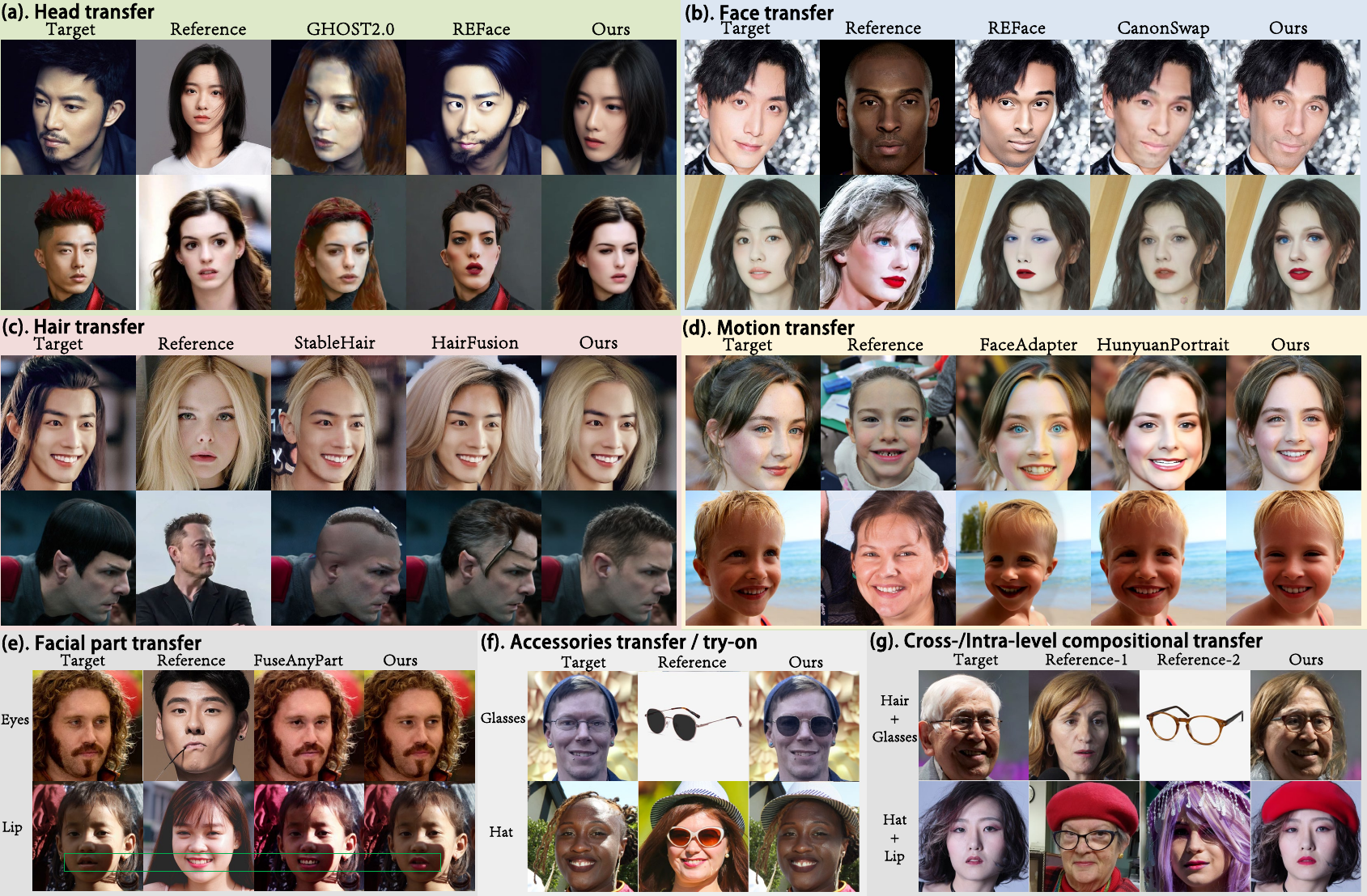}
  \end{center}
  \vspace{-1.5em}
  \caption{Visual Comparisons on diverse deepface tasks. 
More results in Suppl.
}
\vspace{-1.5em}
  \label{fig:comp}
  \label{fig:mainComp}
\end{figure*}

\noindent \textbf{Baselines} \quad 
We compare with methods in three categories, 
1) \textbf{Task-specific approaches} that devotes to solve a specific task, including Face transfer, Motion transfer, Hair transfer, and Head transfer. 
We select recent SOTAs for each task, 
CanonSwap~\cite{canonswap_iccv25} for Face transfer;
HunyuanPortrait~\cite{hunyuanportrait_cvpr25} for Motion transfer;
StableHair~\cite{stable_hair_aaai25}, and HairFusion~\cite{hairfusion_aaai25} for Hair transfer;
and GHOST2.0~\cite{ghost20} for Head transfer.
2) \textbf{Multi-task methods} REFace~\cite{reface_wacv25} that can handle more than 1 task with the same model but need to retrain for different tasks.
3) \textbf{Unified Methods} that can handle more than one task with the same model and only need to train once for all tasks, including Face-Adapter~\cite{faceadapter_eccv24}, and RigFace~\cite{towards_consistent_face_editing_arxiv25}.
Among the diffusion-based baselines, REFace, StableHair, HairFusion, Face-Adapter, and RigFace adopt the same base model as ours (Stable Diffusion v1.5), whereas HunyuanPortrait adopts Stable Video Diffusion.
See Sec. A.5 in Suppl. for more details.

\noindent \textbf{Metrics.} \quad
Following prior works, we use face ID Similarity (ID sim), Pose Distance (pose dist)~\cite{hopenet}, Expression Distance (expr dist), CLIP Distance (CLIP dist), and SSIM.

\noindent \textbf{More Details} can be found in Sec. A of the supplemental material.


\subsection{Evaluations and Discussions}

UniBioTransfer supports a wide range of multi-level biometrics transferring, encompassing three hierarchically organized levels of semantic control. 
Visual Comparisons can be found in Fig.~\ref{fig:comp}, and the supplemental material.


\begin{table*}[t]
    \centering
    \caption{Quantitative comparisons on \textbf{High-Level} and \textbf{Mid-Level} deepface tasks with \textbf{multi-task and unified approaches}.
    Notably, only \textit{UniBioTransfer} can perform all four tasks simultaneously.
    For all metrics, $\uparrow$ indicates higher is better and $\downarrow$ indicates lower is better. 
    Best results are highlighted in \textbf{\color{red}bold}.
    }
    \vspace{-1em}
    \label{tab:swap_comparison-multi}
    \resizebox{\textwidth}{!}{%
    \begin{tabular}{l|ccc|ccc|ccc|ccc}
    \toprule
    \multirow{3}{*}{\textbf{Method}} & 
    \multicolumn{9}{c|}{\textbf{Medium Level Tasks}}  & \multicolumn{3}{c}{\textbf{High Level Tasks}}  \\ 
    \cmidrule(lr){2-10} \cmidrule(lr){11-13}
    &\multicolumn{3}{c|}{\textbf{Face Transfer}} & \multicolumn{3}{c|}{\textbf{Hair Transfer}} & \multicolumn{3}{c|}{\textbf{Motion Transfer}} & \multicolumn{3}{c}{\textbf{Head Transfer}} \\
    & ID sim$\uparrow$ & pose dist.$\downarrow$ & expr. dist.$\downarrow$           & CLIP dist.$\downarrow$ & ID sim$\uparrow$ & non-hair SSIM$\uparrow$               & ID sim$\uparrow$ & pose dist.$\downarrow$ & expr. dist.$\downarrow$         & CLIP dist.$\downarrow$ & ID sim$\uparrow$ & pose dist.$\downarrow$ \\
    \midrule
    \multicolumn{12}{l}{\textbf{Multi-Task Methods (Training separately for different tasks)}} \\
    REFace (WACV 25) & 0.631 & 3.75 & 1.04 & - & - & - & - & - & - & 0.639 & 0.540 & \underline{5.43} \\
    \midrule
    \multicolumn{12}{l}{\textbf{Unified Method (Train once for all tasks)}} \\
    Face-Adapter (ECCV 24) & 0.519 & 4.91 & 1.33 & - & - & - & 0.503 & \textbf{\textcolor{red}{6.89}} & 2.27 & - & - & - \\
    RigFace (arXiv 25) & 0.357 & 5.46 & 2.11 & - & - & - & 0.413 & 9.33 & 2.32 & - & - & - \\
    \textbf{UniBioTransfer(Ours)} & \textbf{\textcolor{red}{0.637}} & \textbf{\textcolor{red}{3.63}} & \textbf{\textcolor{red}{1.03}}      & \textbf{\textcolor{red}{0.421}} & \textbf{\textcolor{red}{0.887}} & \textbf{\textcolor{red}{0.91}}     & \textbf{\textcolor{red}{0.602}} & 7.09 & \textbf{\textcolor{red}{2.09}}       & \textbf{\textcolor{red}{0.460}} & \textbf{\textcolor{red}{0.545}} & \textbf{\textcolor{red}{5.28}} \\
    \bottomrule
    \end{tabular}%
    }
    \end{table*}
    

\begin{table*}[t]
    \centering
    \caption{Quantitative comparisons on \textbf{High-Level} and \textbf{Mid-Level} deepface tasks with \textbf{task-specific methods}. 
    Notably, \textit{UniBioTransfer} performs all the four tasks simultaneously in a single training cycle.
    }
    \vspace{-1em}
    \label{tab:swap_comparison-single}
    \resizebox{\textwidth}{!}{%
    \begin{tabular}{l|ccc|ccc|ccc|ccc}
    \toprule
    \multirow{3}{*}{\textbf{Method}} & 
    \multicolumn{9}{c|}{\textbf{Medium Level Tasks}}  & \multicolumn{3}{c}{\textbf{High Level Tasks}}  \\ 
    \cmidrule(lr){2-10} \cmidrule(lr){11-13}
    &\multicolumn{3}{c|}{\textbf{Face Transfer}} & \multicolumn{3}{c|}{\textbf{Hair Transfer}} & \multicolumn{3}{c|}{\textbf{Motion Transfer}} & \multicolumn{3}{c}{\textbf{Head Transfer}} \\
    & ID sim$\uparrow$ & pose dist.$\downarrow$ & expr. dist.$\downarrow$           & CLIP dist.$\downarrow$ & ID sim$\uparrow$ & non-hair SSIM$\uparrow$               & ID sim$\uparrow$ & pose dist.$\downarrow$ & expr. dist.$\downarrow$         & CLIP dist.$\downarrow$ & ID sim$\uparrow$ & pose dist.$\downarrow$ \\
    \midrule
    HunyuanPortrait (CVPR 25) & - & - & - & - & - & - & 0.580 & 9.22 & 2.18 & - & - & - \\
    CanonSwap (ICCV 25) & 0.551 & \textbf{\textcolor{red}{2.92}} & \textbf{\textcolor{red}{0.87}} & - & - & - & - & - & - & - & - & - \\
    Stable-Hair (AAAI 25) & - & - & - & 0.468 & 0.855 & 0.87 & - & - & - & - & - & - \\
    HairFusion (AAAI 25) & - & - & - & 0.478 & 0.848 & 0.80 & - & - & - & - & - & - \\
    GHOST2.0 (arXiv 25) & - & - & - & - & - & - & - & - & - & 0.618 & 0.461 & 5.78 \\
    \midrule
    \textbf{UniBioTransfer(Ours)} & \textbf{\textcolor{red}{0.637}} & \textcolor{black}{3.63} & {\textcolor{black}{1.03}}      & \textbf{\textcolor{red}{0.421}} & \textbf{\textcolor{red}{0.887}} & \textbf{\textcolor{red}{0.91}}     & \textbf{\textcolor{red}{0.602}} & \textbf{\textcolor{red}{7.09}} & \textbf{\textcolor{red}{2.09}}       & \textbf{\textcolor{red}{0.460}} & \textbf{\textcolor{red}{0.545}} & \textbf{\textcolor{red}{5.28}} \\
    \bottomrule
    \end{tabular}%
    }
    \end{table*}
    
\begin{table*}[!htb]
    \caption{Quantitative comparisons on \textbf{Low-Level} Facial-Part Swap tasks. 
}
     \vspace{-1em}
    \label{tab:swap_comparison-facial-part}
    \centering
    \setlength{\tabcolsep}{9.8pt}
    \resizebox{0.99\textwidth}{!}{%
    \begin{tabular}{l|ccc|ccc|ccc}
    \hline
    \multirow{2}{*}{\textbf{Method}} & \multicolumn{9}{c}{\textbf{Facial Part Transfer (Low Level Tasks)}} \\ \cline{2-10}
                                    & \multicolumn{3}{c|}{\textbf{Eye}} & \multicolumn{3}{c|}{\textbf{Nose}} & \multicolumn{3}{c}{\textbf{Lip}} \\
                                    & CLIP L2$\downarrow$ & expr L2$\downarrow$ & SSIM$\uparrow$ & CLIP L2$\downarrow$ & expr L2$\downarrow$ & SSIM$\uparrow$ & CLIP L2$\downarrow$ & expr L2$\downarrow$ & SSIM$\uparrow$ \\ \hline
    FuseAnyPart (NIPS 24)                      & \underline{0.449} & 0.798 & \underline{0.898} & \underline{0.332} & 0.802 & \underline{0.898} & \textbf{\color{red}0.372} & \underline{0.794} & \underline{0.902} \\
    Ours (w/o Uni.)                      & 0.471 & \underline{0.697} & 0.889 & 0.337 & \underline{0.672} & 0.859 & 0.384 & 1.009 & 0.867 \\
    \textbf{Ours}                    & \textbf{\color{red}0.427} & \textbf{\color{red}0.599} & \textbf{\color{red}0.927} & \textbf{\color{red}0.327} & \textbf{\color{red}0.532} & \textbf{\color{red}0.937} & \underline{0.378} & \textbf{\color{red}0.579} & \textbf{\color{red}0.937} \\ \hline
    \end{tabular}
    }
    \end{table*}

\noindent \textbf{High-level Head Transfer} \quad
First, as shown in Table~\ref{tab:swap_comparison-multi}, no other unified baselines can handle the high-level head transfer task due to the task complexity.
Meanwhile, our unified solution also significantly outperforms SoTA task-specific  Head transfer approaches (Table~\ref{tab:swap_comparison-single}), and retraining-required multi-task method REFace (Table~\ref{tab:swap_comparison-multi}). 

\noindent \textbf{Medium-Level Facial and Hair Transfer} \quad
Meanwhile, \textit{UniBioTransfer} is the only unified model that can solve 3 medium-level tasks at the same time with the high-level head transfer task within a single training cycle, including Face transfer, Face Reenactment, and Hair Transfer (Table ~\ref{tab:swap_comparison-multi} and~\ref{tab:swap_comparison-single}). 

\noindent \textit{Hair Transfer} \quad 
It is worth noting that since hair is extremely complicated compared to face, in both structure and color variations, neither existing multi-task nor unified methods can handle the hair transfer task (Table~\ref{tab:swap_comparison-multi}).
To the best of our knowledge, we are the first to include the hair transfer task in a unified deepface solution.
Note that we also outperform task-specific Hair transfer approaches, Stable Hair and Hair Fusion (Table~\ref{tab:swap_comparison-single}). 

\noindent \textit{Face and Motion Transfer} \quad
We also demonstrate superior performance over all multi-task and unified methods across most metrics on the two tasks (Table~\ref{tab:swap_comparison-multi}), while achieving competitive or even improved results compared to task-specific baselines (Table~\ref{tab:swap_comparison-single}).

\noindent \textit{Low-Level Fine-grained Feature Transfer} \quad
Starting from our pre-trained multi-task model, new low-level tasks (e.g., eye/lip transfer) are adapted by fine-tuning on $<20\%$ of data and $<10\%$ training cost relative to its inherited task (face transfer). Initialization is created by copying all task-specific parameters from the inherited task, including task-specific experts, router, encoding projections, and attention projections.
As shown in Table~\ref{tab:swap_comparison-facial-part}, we achieve superior performance to the SOTA specialized method FuseAnyPart~\cite{fuseanypart_neurips24}. 
As illustrated in green box in Fig.~\ref{fig:mainComp} (e) and discussed in the limitation section of their paper~\cite{fuseanypart_neurips24}, limited expression preservation is an inherent drawback of FuseAnyPart. 
Its better lip CLIP metric than ours partially comes from its tendency to directly copy-and-paste the mouth without preserving the target’s expression and skin tone.

We also include a variant that is trained from the base model instead of our pre-trained unified model, with $\sim$2$\times$ training cost, denoted as \textit{Ours (w/o Uni.)} in Tab.~\ref{tab:swap_comparison-facial-part}, which still underperforms our unified-pretrained model and thus shows the generalization ability benefit from the unification.

\noindent \textbf{Cross-Level and Intra-Level Compositional Transfer} \quad
Our UniBioTransfer can also be fast adapted to compositional tasks (transferring combinations of attributes), as illustrated in Fig.~\ref{fig:teaser}.
Specifically, for the initialization of a compositional task, we inherit two task-specific experts from the corresponding parent tasks and introduce a dedicated router to balance their contributions for each token.
This flexibility highlights our framework's core advantage: a shared knowledge base that enables conflict-free task combinations.


\vspace{-0.2em}
\subsection{Ablation Studies}
\vspace{-0.2em}
\label{sec:ablation}

\vspace{-3.2em}
\begin{figure}[H]
  \centering
  \begin{minipage}[t]{0.49\linewidth}
    \centering
    \captionof{table}{Ablation studies on model design and training strategy. For each task, we report its most representative metric (identity similarity, hair clip distance, pose distance, and head clip distance respectively).}
    \label{tab:ablation_moe}
    \vspace{-0.18em}

\resizebox{\linewidth}{!}{
\begin{tabular}{@{}lcccc@{}}
\toprule
\multicolumn{5}{c}{\textbf{Model Design}} \\ \midrule
\multicolumn{1}{l|}{\textbf{Setting}} &
  \textbf{Face $\uparrow$} &
  \textbf{Hair $\downarrow$} &
  \textbf{Motion $\downarrow$} &
  \textbf{Head $\downarrow$} \\ \midrule
\multicolumn{1}{l|}{w/o BioMoE (Naive Parameter Sharing)} &
  0.478 &
  0.474 &
  9.74 &
  \multicolumn{1}{c|}{0.574} \\
\multicolumn{1}{l|}{w/o BioMoE (Task-Specific Model)} &
  0.622 &
  0.428 &
  7.14 &
  \multicolumn{1}{c|}{0.465} \\
\multicolumn{1}{l|}{w/o Structure-Aware Routing} &
  0.634 &
  0.423 &
  7.13 &
  \multicolumn{1}{c|}{0.469} \\
\multicolumn{1}{l|}{\textbf{Ours}} &
  \textbf{0.637} &
  \textbf{0.421} &
  \textbf{7.09} &
  \textbf{0.460} \\ \midrule
\multicolumn{5}{c}{\textbf{Training Strategy}} \\ \midrule
\multicolumn{1}{l|}{\textbf{Setting}} &
  \textbf{Face $\uparrow$} &
  \textbf{Hair $\downarrow$} &
  \textbf{Motion$\downarrow$} &
  \textbf{Head $\downarrow$} \\ \midrule
\multicolumn{1}{l|}{w/o Two-Stage Training} &
  0.507 &
  0.443 &
  8.33 &
  \multicolumn{1}{c|}{0.491} \\
\multicolumn{1}{l|}{w/o Result-based Rank Allocation} &
  0.634 &
  0.433 &
  7.35 &
  \multicolumn{1}{c|}{0.464} \\
\multicolumn{1}{l|}{\textbf{Ours}} &
  \textbf{0.637} &
  \textbf{0.421} &
  \textbf{7.09} &
  \multicolumn{1}{c|}{\textbf{0.460}} \\ \bottomrule
\end{tabular}
}
  \end{minipage}\hfill
  \begin{minipage}[t]{0.49\linewidth}
    \centering
    \captionof{table}{Ablations on data corruption strategy. CLIP distance ($\downarrow$) is used to evaluate the ability to transfer hair/head.}
    \label{tab:ablation_data}
    \vspace{-0.2em}

\resizebox{\linewidth}{!}{%
\begin{tabular}{l|cc}
\toprule
\textbf{Method} & \textbf{Hair Transfer $\downarrow$ } & \textbf{Head Transfer $\downarrow$}  \\
\midrule
w/o swapping-based corruption & 0.492 & 0.555  \\
\textbf{Ours} & \textbf{0.421} & \textbf{0.460} \\
\bottomrule
\end{tabular}%
}

    \vspace{0.6em}
    \includegraphics[width=\linewidth]{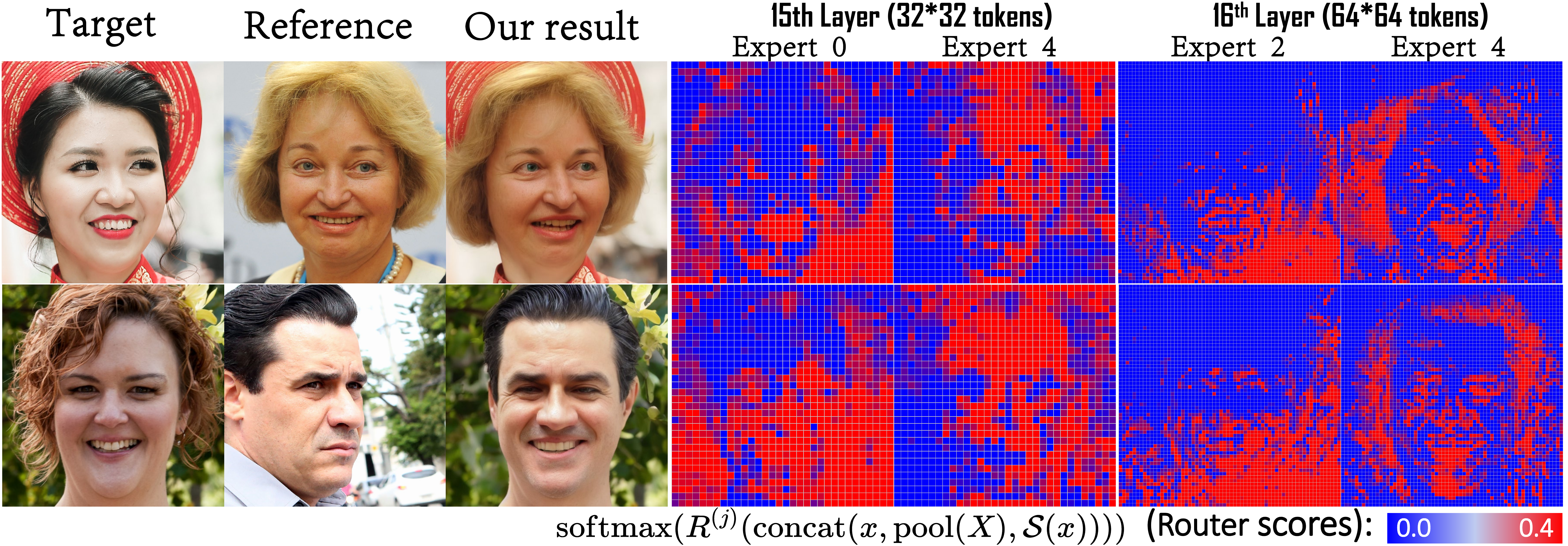}
    \vspace{-2.0em}
\captionof{figure}{
Structure-aware routing scores after softmax and before top-$K$ selection.
}
    \label{fig:model_routing}
  \end{minipage}
\end{figure}

\vspace{-0.3em}
\textbf{\textit{Ablation on Model Design.}} \quad
To validate our BioMoE, we compare against two variations in Table~\ref{tab:ablation_moe}: 1) \textbf{Single-Task Models}, where we train a separate model for each task, and 2) \textbf{Naive Sharing}, where a single model with fully shared parameters is trained on all tasks.
The consistently inferior performance of Naive Sharing highlights the severity of cross-task conflicts and motivates our BioMoE design.
Our unified training not only yields competitive performance against task-specific models across tasks, but also offers practical advantages in deployment by replacing multiple specialized models with a single checkpoint. In addition, the trained unified model serves as a strong starting point for fast adaptation to new tasks (including low-level transfers and cross-/intra-level compositional transfer) with limited additional data and computation, which is quantitatively validated by the superior  performance of our model compared to the 'w/o Uni.' baseline in Table~\ref{tab:swap_comparison-facial-part} (which adopts task-specific training at $\sim$2$\times$ the training cost).

We then compare our \textbf{Structure-Aware Routing} to the vanilla router that relies only on input tokens. 
In Fig.~\ref{fig:model_routing}, we visualize expert average probabilities during head transfer inference,
indicating that experts specialize to semantically structured token regions.
For instance, Layer 15's Expert 4 focuses more on the forehead, and Layer 16's Expert 2 focuses on the general neck region, likely handling the seamless blending of the transferred head with the target neck and background.

\textbf{\textit{Ablation on Training Strategy.}} \quad
In Table~\ref{tab:ablation_moe},  we also validate our proposed Two-Stage Training strategy by comparing against a one-stage baseline that trains the MoE model jointly from the start.
We also show that our Result-based Adaptive LoRA Rank Allocation performs better than a vanilla gradient-based one (w/o Result-based Rank Allocation). Additional comparisons are detailed in Sec. B.2 in the supplementary material.


\textbf{\textit{Ablation on Data Construction Solution.}} \quad
We analyze the impact of our proposed \textit{Swapping-Based Corruption} for spatially-dynamic tasks (Hair and Head transfer).  
In Table~\ref{tab:ablation_data}, we compare against a baseline model where the data for these tasks is constructed using a simple masking approach, confirming that our data solution to prevent the model from accessing trivial solutions is critical.

\vspace{-0.1em}
\section{Conclusions}
\label{sec:conclusion}
\vspace{-0.1em}
To conclude, we present \textbf{UniBioTransfer}, a unified, multi-task framework for diverse DeepFace generation tasks. Leveraging our unified data construction strategy and an innovative BioMoE component with a two-stage training strategy, the framework effectively handles shape variations, mitigates task interference, and exploits task synergy, significantly outperforming state-of-the-art unified models while maintaining competitive or superior performance compared to task-specific models. Importantly, \textbf{UniBioTransfer} can be efficiently adapted to new tasks with minimal fine-tuning, demonstrating strong generalization capability and scalability.


\clearpage
\appendix
\renewcommand{\theHsection}{\Alph{section}}
\renewcommand{\theHsubsection}{\Alph{section}.\arabic{subsection}}
The supplementary material is organized as follows:
\begin{itemize}
    \item Section~\ref{sec:suppl_impl} provides comprehensive implementation details:
    \begin{itemize}
        \item Sec.~\ref{sec:suppl_model_train} \& \ref{sec:suppl_loss}: The model pipeline, model training process, and algorithms for initialization and forwarding.
        \item Sec.~\ref{sec:suppl_data}: Data construction strategies.
        \item Sec.~\ref{sec:suppl_eval} \& \ref{sec:impl_sota}: Evaluation protocols and baseline settings.
        \item Sec.~\ref{sec:suppl_adapt}: Adaptation to new tasks.
    \end{itemize}

    \item Section~\ref{sec:suppl_analysis} presents extensive analysis and ablation studies:
    \begin{itemize}
        \item Sec.~\ref{sec:suppl_gen_model} discusses the generative models used in data corruption.
        \item Sec.~\ref{sec:suppl_abla_lora}: Quantitative ablations on adaptive LoRA rank allocation.
        \item Sec.~\ref{sec:suppl_data_filter}: Visual examples of our data filtering.
        \item Sec.~\ref{sec:suppl_stylegan} discusses StyleGAN-based methods.
    \end{itemize}

    \item Section~\ref{sec:suppl_quant} provides extended quantitative comparisons against state-of-the-art methods:
    \begin{itemize}
        \item Sec.~\ref{sec:suppl_metrics}: Additional metrics.
        \item Sec.~\ref{sec:suppl_efficiency}: A computational efficiency analysis.
        \item Sec.~\ref{sec:suppl_complex_scenes}: Evaluations under challenging conditions like extreme poses, exaggerated expressions, and occlusions.
    \end{itemize}

    \item Section~\ref{sec:suppl_vis} provides additional visual comparisons across various tasks.
    
    \item Section~\ref{sec:suppl_limit} discusses the limitations of our work.
\end{itemize}

\section{Implementation Details}
\label{sec:suppl_impl}


\subsection{Model and training process}
\label{sec:suppl_model_train}

\textbf{Overall Pipeline} \quad
Our framework is built upon the Stable Diffusion v1.5\cite{stableDiffusion} backbone.
As shown in Fig. 4 in the main paper, 
given a target image $I_\text{tgt}$ and a reference image $I_\text{ref}$, we first mask out irrelevant attributes in both.  
Following REFace~\cite{reface_wacv25}, we extract semantic features of masked $I_\text{ref}$ via the CLIP vision encoder~\cite{openai_clip} and identity features (for tasks that $I_\text{ref}$ provides identity attribute like face transfer and head transfer) via the Arcface face recognizer~\cite{deng2019arcface}. 
Task-specific single-layer MLPs project these raw features into tokens  for cross attention conditioning in Unets.  
We form blended landmarks to guide motion, taking pose/expression from the original driving image ($I_\text{ref}$ for reenactment, $I_\text{tgt}$ otherwise) and identity from the counterpart.  
The masked target is encoded by the VAE to a latent and concatenated with noise along the channel dimension as input to the main UNet.  
The masked reference flows through the refNet (Reference-UNet), and its tokens are injected via cross-attention into the main UNet (the context tokens in main Unet are concatenated with the tokens from refNet).
Different from common refNet used in other works~\cite{animate_anyone,xu2024magicanimate,wei2024aniportrait,stable_hair_aaai25,hairfusion_aaai25,reface_wacv25,faceadapter_eccv24}, we use only the first half of the refNet as we found it is enough to achieve a good performance and yields better efficiency.
Finally, the main UNet performs iterative denoising on the concatenated latents, which are then decoded back into pixel space by the VAE decoder to produce the final swapped result $I_\text{out}$.

\textbf{Training process} \quad

In the main paper for simplicity, we only present the MoE mechanism for FFN. Here we show a more detailed MOE mechanism. 
In addition to modifying the FFN layers, we also adapt the attention mechanism to be task-aware to further mitigate task interference. Specifically, we make the query (Q), key (K), value (V), and output projection matrices task-specific.
By creating separate projections for each task, the self-attention layer can generate distinct feature representations tailored to the specific requirements of each task before the features are passed to the subsequent BioMoE layer. 
This preemptive feature differentiation helps to alleviate potential gradient conflicts that could arise if all experts were to operate on an identical input feature space.
This design is parameter-efficient, as the attention projections are a very small fraction of the overall model size.

In stage one, we isolate gradients across tasks for attention and FFN. After training for 100K steps, we proceed to stage-II. 
Initialization of stage-II:

1) Weight initialization of stage II.
As in Section 3.3 of the main paper, we average Stage-I FFNs to initialize the shared/global expert: $W_{\text{gl}}=\frac{1}{N}\sum_{j}W_j$. For task $j$, its task-specific expert is initialized on the residual $W_j-W_{\text{gl}}$. Note that each specialized expert is actually an FFN with 2 MLPs in LoRA form (but we depicted it as a single LoRA structure for simplicity in the algorithm part and figure). We perform SVD on the residual and select the smallest rank $r$ such that the cumulative energy exceeds a global threshold $\tau=0.2$, yielding an adaptive yet unified criterion across tasks/layers.

\noindent\textbf{Pseudocode: Weight initialization of stage II}
\begin{enumerate}
  \item Inputs: Stage-I FFNs $\{W_j\}_{j=1}^N$, global threshold $\tau$.
  \item Global: $W_{\text{gl}} \leftarrow \frac{1}{N}\sum_{j}W_j$.
  \item For each task $j$: SVD $(W_j-W_{\text{gl}})=U\Sigma V^\top$.
  \item Pick $r$ s.t. $\sum_{i\le r}\sigma_i^2/\sum_i\sigma_i^2 \ge \tau$ (same $\tau$ for all).
  \item Initialize task-$j$ LoRA by rank-$r$ reconstruction of the residual.
\end{enumerate}

\begin{algorithm}[t]
\caption{Weight initialization of stage II}
\label{alg:init}
\small
\begin{algorithmic}[1]
\Require 
    set of $N$ trained task-specific FFN weights $\{W_1, \dots, W_N\}$, 
    energy threshold $\tau$.
\Ensure 
    Initialized Global Expert $W_{\text{gl}}$, 
    set of task-specific experts $\{ (A_j, B_j) \}_{j=1}^N$.

\Statex \textbf{Declarations:}
\Statex \quad $j \in \{1, \dots, N\}$ \Comment{Task ID}
\Statex \quad $W_j, \bar{W}, W_{\text{gl}} \in \mathbb{R}^{d_{\text{out}} \times d_{\text{in}}}$ \Comment{FFN weight matrices}
\Statex \quad $S \in \mathbb{R}^{\min(d_{\text{out}}, d_{\text{in}})}$ \Comment{Singular values}

\State \textbf{Step 1: Consensus Initialization}
\State $\bar{W} \gets \frac{1}{N} \sum_{j=1}^{N} W_j$ \Comment{Compute consensus (average) weights}
\State $W_{\text{gl}} \gets \bar{W}$ \Comment{Initialize Global Expert}

\State \textbf{Step 2: Residual-based Adaptive LoRA Initialization}
\For{$j = 1$ to $N$}
    \State $U, S, V^\top \gets \text{SVD}(W_j - W_{\text{gl}})$ \Comment{Singular Value Decomposition of residual}
    \State $E_{\text{total}} \gets \sum_{k} S_k^2$ \Comment{Calculate total energy}
    \State $r_j \gets 0, \quad E_{\text{cum}} \gets 0$
    \While{$E_{\text{cum}} / E_{\text{total}} < \tau$} \Comment{Determine adaptive rank}
        \State $r_j \gets r_j + 1$
        \State $E_{\text{cum}} \gets E_{\text{cum}} + S_{r_j}^2$
    \EndWhile
    \State Initialize LoRA parameters $A_j, B_j$ with rank $r_j$ to approximate the residual.
\EndFor
\end{algorithmic}
\end{algorithm}

2) In Stage~II, we introduce $E{=}8$ routed experts and a per-task router $R^{(j)}$ (single-layer MLP). All routed experts share the FFN topology but reduce the hidden width by factor $F{=}16$, balancing capacity and cost. Parameters are randomly initialized before joint training.

After initialization of stage II, we jointly train the N=4 tasks (including face transfer, hair transfer, motion transfer, and head transfer). We train for 120K steps. 

\begin{algorithm}[t]
\caption{BioMoE Forward Process}
\label{alg:moe_forward}
\small
\begin{algorithmic}[1]
\Require 
    Input token feature $x \in \mathbb{R}^{D}$, 
    Global context feature $\operatorname{pool}(X) \in \mathbb{R}^{D}$,
    Structure information $\mathcal{S}(x) \in \mathbb{R}^{2M}$,
    Task ID $j$.
\Ensure 
    Output feature $f(x) \in \mathbb{R}^{D}$.

\Statex \textbf{Declarations:}
\Statex \quad $f_{\mathrm{gl}}: \mathbb{R}^D \to \mathbb{R}^D$ \Comment{Global Shared Expert}
\Statex \quad $\{f_{\mathrm{rt}}^{(i)}\}_{i=1}^E, f_{\mathrm{rt}}^{(i)}: \mathbb{R}^D \to \mathbb{R}^D$ \Comment{Set of $E$ Routed Experts}
\Statex \quad $f_{\mathrm{sp}}^{(j)}: \mathbb{R}^D \to \mathbb{R}^D$ \Comment{Task-specific Expert of task-$j$.}
\Statex \quad $R^{(j)}: \mathbb{R}^{D_{\text{gate}}} \to \mathbb{R}^E$ \Comment{Task-$j$ Structure-Aware Router}
\Statex \quad $K \in \mathbb{Z}^+$ \Comment{Number of active experts}
\Statex \quad $\epsilon \sim \mathcal{N}(0, 1)$ \Comment{Gaussian noise for exploration}

\State \textbf{Step 1: Structure-Aware Routing}
\State $g \gets \operatorname{concat}(x, \operatorname{pool}(X), \mathcal{S}(x))$ \Comment{Assemble gating input}
\State $s \gets R^{(j)}(g)$ \Comment{Predict expert scores}
\State $\mathcal{G}^{(j)}(x) \gets \mathrm{top}_K(\mathrm{softmax}(s + \epsilon))$ \Comment{Select top-$K$ experts}

\State \textbf{Step 2: Aggregate Experts and Final Fusion}
\State $f(x) \gets f_{\mathrm{gl}}(x) + f_{\mathrm{sp}}^{(j)}(x) + \sum_{i=1}^{E} \mathcal{G}^{(j)}_i(x) \cdot f_{\mathrm{rt}}^{(i)}(x)$ 
\end{algorithmic}
\end{algorithm}

The model is trained on 4 NVIDIA A40 GPUs for around 8.5 days. Each GPU receives data from one task and the batch size is set to one. 
In Stage~I, each task is optimized separately without gradient synchronization. In Stage~II, we start from the Weight initialization of stage II and perform joint training with synchronous updates at each iteration.

\textbf{We will release the source code and model weight} which contains more details for better reproducibility. 

\subsection{Training loss}
\label{sec:suppl_loss}
Following REFace~\cite{reface_wacv25}, our training objective consists of a standard DDPM loss $\mathcal{L}_{\text{diff}}$ and a multi-step DDIM-based perceptual loss $\mathcal{L}_{\text{perceptual}}$:
\begin{equation}
  \mathcal{L} = \mathcal{L}_{\text{diff}} + \lambda_{\text{perceptual}} \mathcal{L}_{\text{perceptual}}(I_{\text{gt}}, \hat{I}) + \lambda_{\text{id}} \mathcal{L}_{\text{id}}(I_{\text{gt}}, \hat{I})
  \label{eq:loss}
\end{equation}
where $\mathcal{L}_{\text{diff}}$ is the standard noise prediction loss at latent space, and $\mathcal{L}_{\text{perceptual}}, \mathcal{L}_{\text{id}}$ are image-space enhancement losses computed between DDIM-sampled image $\hat{I}$ and the ground-truth $I_{\text{gt}}$ at pixel space.

\noindent\textbf{Latent-space Loss.} We utilize the standard noise prediction loss $\mathcal{L}_{\text{diff}}$ to train the diffusion model:
\begin{equation}
    \mathcal{L}_{\text{diff}} = \mathbb{E}_{z, \epsilon \sim \mathcal{N}(0,1), t} [ \| \epsilon - \epsilon_\theta(z_t, t, \mathbf{c}) \|^2_2 ],
\end{equation}
where $z_t$ is the noisy latent at timestep $t$ and $\mathbf{c}$ represents the task-specific conditions, including the reference features and structure guidance. This loss allows the model to efficiently learn the mapping for various biometric transfer tasks in the compressed latent space.

\noindent\textbf{Image-space Loss.} To further refine the visual details and enhance identity preservation, we incorporate image-space losses computed on the reconstructed image $\hat{I}$:
\begin{equation}
    \mathcal{L}_{\text{img}} = \lambda_{\text{perceptual}} \mathcal{L}_{\text{perceptual}}(I_{\text{gt}}, \hat{I}) + \lambda_{\text{id}} \mathcal{L}_{\text{id}}(I_{\text{gt}}, \hat{I}).
\end{equation}
The reconstructed image $\hat{I}$ is obtained by applying $N_{\text{step}}=4$ DDIM denoising steps starting from a noisy latent. Each DDIM step requires a complete feed-forward pass of the network. Consequently, the memory requirement for computing $\mathcal{L}_{\text{img}}$ is significantly higher than that for $\mathcal{L}_{\text{diff}}$, as it involves $N_{\text{step}}$ times the computation and gradient storage. This leads to a reduced batch size and slower training speed when these losses are active.

\noindent\textbf{Training Schedule.} Given the computational overhead of the image-space losses, we adopt a two-phase weighting strategy. To distinguish these from the primary Stage-I and Stage-II of our Two-Stage Training Strategy, we refer to them as \textit{sub-stages}. During the first sub-stage (the majority of training), we set $\lambda_{\text{perceptual}} = \lambda_{\text{id}} = 0$, focusing on $\mathcal{L}_{\text{diff}}$ to achieve fast convergence in terms of structure and basic attribute transfer. In the final sub-stage, we fine-tune the model with $\lambda_{\text{perceptual}} > 0$ and $\lambda_{\text{id}} > 0$ for a few epochs to boost perceptual realism and identity fidelity.

\subsection{Training Data}
\label{sec:suppl_data}
\subsubsection{Datasets}
To acquire real images and videos for training data construction, for the face transfer task, we utilize 
the entire VFHQ~\cite{xie2022vfhq} dataset, which consists of 15K videos 
and a subset of the Arc2Face dataset~\cite{paraperas2024arc2face}, comprising 60K images across 30K distinct identities.
For glasses tasks, we use dataset from~\cite{kaggle_glasses_dataset} containing 7K real images with glasses after filter. For hat tasks we use images with hats in FFHQ~\cite{ffhq} (separate from testset) and VFHQ~\cite{xie2022vfhq} dataset, resulting in 5K real images. For other tasks, celebV-HQ~\cite{zhu2022celebV_HQ} and VFHQ~\cite{xie2022vfhq} are used. After filter, we get 80K real images from the two datasets. For each real image, we will generate one to two training pairs. 

\subsubsection{Training data construction}

Our unified data corruption strategy handles different attribute types, 
(a) Relative-static attributes. 
Relative-static attributes include both structural and non-structural elements whose spatial locations and overall shapes remain largely consistent during transfer.
For structural attributes, e.g., face, eyes, lips, nose, which have well-defined geometric boundaries, we apply mask-based corruption, a widely used and effective strategy that removes texture and color information, forcing the model to reconstruct these details from the reference.
For non-structural attributes, e.g., skin tone, we employ standard data augmentation techniques to introduce variation while preserving structure.
Taking face transfer as an example, the target image $I'_{\text{tgt}}$ is constructed by masking out the face in $I_{\text{gt}}$, preserving only the background, skin tone (from the neck region), and pose-expression (from landmarks). The reference image $I'_{\text{ref}}$ is created by corrupting all attributes except the face: background, hair, skin tone (via lighting augmentation), and pose-expression (via warp augmentation or selecting a different frame from the video).
(b) Spatially-dynamic attributes.
Taking hair transfer as an example, the target image $I'_{\text{tgt}}$ is constructed by corrupting the hair in $I_{\text{gt}}$ using the swapping-based corruption strategy.

\begin{figure}[htbp]
  \begin{center}
    \includegraphics[width=  0.73  \textwidth]{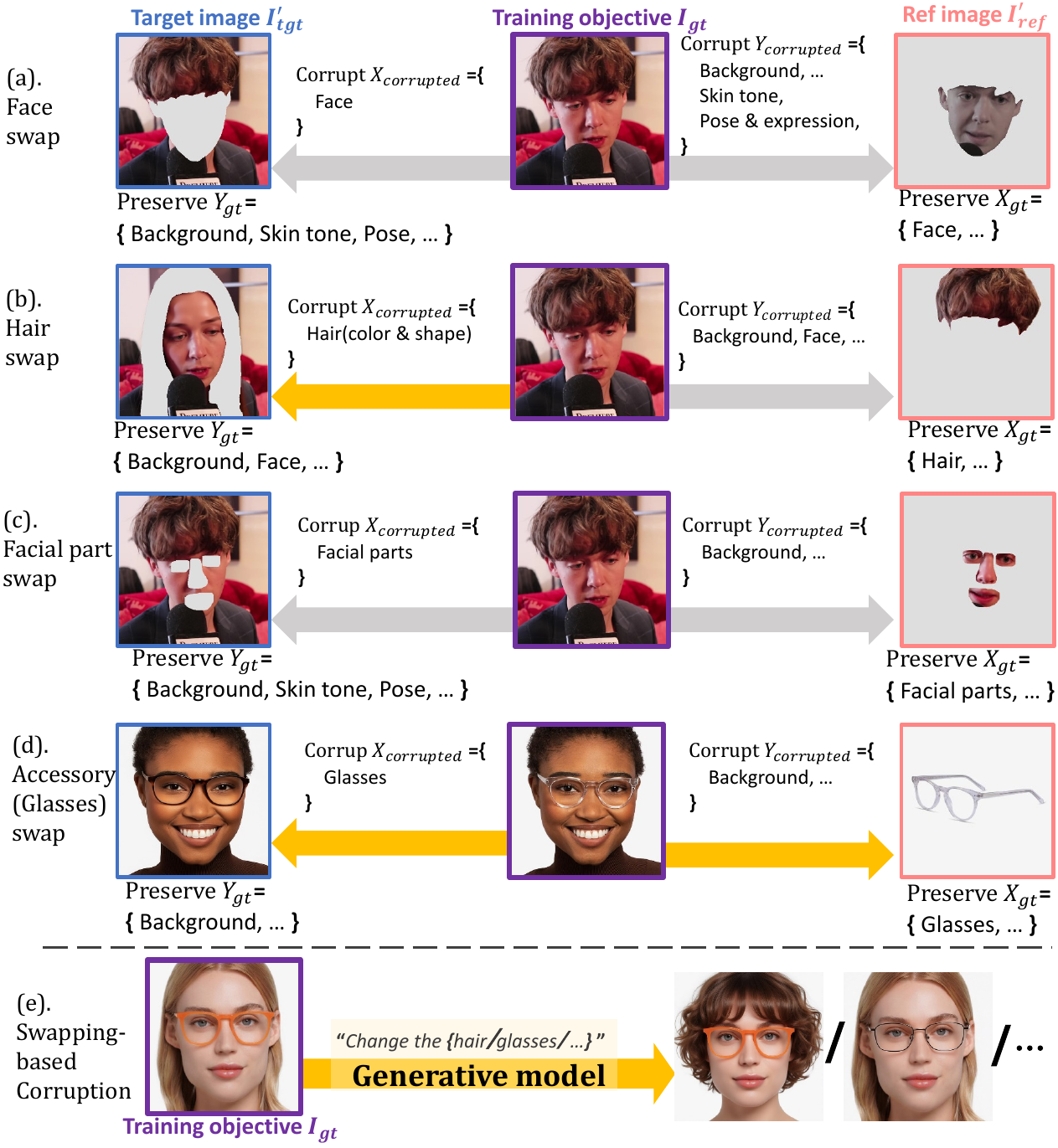}
  \end{center}
  \caption{We show the specific examples of data construction on 4 tasks in (a)-(d). 
Grey arrow means the corruption process only involves Relative-Static Attributes Corruption.
Yellow arrow means it involves Spatially-Dynamic Attributes Corruption, thus need the Swapping-Based Corruption, as shown in (e).
}\label{fig:more-data_cons}
\end{figure}

In Fig~\ref{fig:more-data_cons}, we show four specific examples of data construction.

\subsubsection{Preprocess}
All inputs are resized to $512 \times 512$, and irrelevant regions are masked out for each image (e.g., for head transfer we preserve only the head region in the reference image and mask out the head region in the target image) to ensure consistency with the training process.
We use the semantic mask extracted from SegNeXt-FaceParser~\cite{SegNeXt-FaceParser,e4s} following previous work~\cite{reface_wacv25,faceadapter_eccv24,towards_consistent_face_editing_arxiv25}.
The network outputs at $512 \times 512$ resolution.

For landmark detection, we utilize the MediaPipe model, specifically using the facial part landmarks including pupil points while discarding all face contour points. 
For face-related tasks, when the facial structures (measured by nose-to-mouth distance, nose-to-eye distance, and inter-pupil distance) of the two input images differ substantially, we employ a rule-based landmark blending strategy to combine the identity from one landmark set with the motion from the other, so that the blended landmarks can better guide both identity preservation and motion transfer.
Given two MediaPipe 3D landmark sets, A (expression source) and B (identity target), we compute blended landmarks by preserving B’s facial proportions and injecting A’s expression. We estimate a canonical vertical facial direction by fitting a line to the eye midpoint, nose tip, and mouth center. Then we translate the mouth region and the eye/eyebrow region of A along this vertical direction such that the projected nose-to-mouth and nose-to-eye distances match those measured on B. Finally, we enforce the inter-pupil distance of B through symmetric horizontal translations of the left/right eye regions, while leaving other landmarks unchanged. The blended 3D landmarks are finally projected into 2D landmarks as output.

\subsection{Evaluation Protocol}
\label{sec:suppl_eval}
Following the evaluation protocol established in previous work~\cite{reface_wacv25}, we construct our test set using 1,000 image pairs sampled from the FFHQ dataset. To further ensure a fair comparison with other SoTA methods that rely on some pre-trained detection models (e.g., dlib), we exclude 13 pairs where these detectors completely fail. Consequently, our final evaluation set consists of 987 image pairs.

\subsection{Evaluation of Baselines}
\label{sec:impl_sota}
For most state-of-the-art (SOTA) methods, we utilize their official codebases to perform inference on the test set.

\noindent\textbf{RigFace.} Due to its reliance on multiple pre-trained detection models, RigFace is sensitive to failure cases. During metric calculation, we exclude failed pairs and compute metrics only on successful samples. 

\noindent\textbf{REFace for head transfer.} REFace supports both face and head transfer via mask shuffling as mentioned in its paper. As the official release only includes training code for face transfer, we modify the training code to support head transfer by expanding the mask region to encompass the hair and accessories, following the methodology described in its paper. We then retrain the model using the official hyperparameters.

\noindent\textbf{GHOST 2.0.} For the head transfer task, we follow the setting of previous work~\cite{zeroshot_headswap_cvpr25}, where the goal is to transfer the head along with the skin tone of the reference. GHOST 2.0~\cite{ghost20} adopts a different setting that preserves the target's skin tone. Therefore, in our visual comparisons, we should only focus on the structural alignment and identity transfer quality rather than skin tone consistency for this specific baseline. 

\noindent\textbf{FaceAdapter.} For the face transfer task, we follow the setting of previous works~\cite{reface_wacv25,canonswap_iccv25}, where the face shape in the target image should be preserved. FaceAdapter~\cite{faceadapter_eccv24} adopts a different setting that transfers the reference's face shape. Therefore, in our visual comparisons, we should ignore the face shape preservation for this specific baseline.

\subsection{Adaptation to New Tasks}
\label{sec:suppl_adapt}
\noindent\textbf{Handling Multiple References.} To address compositional tasks that involve multiple reference images (e.g., combining hair from one image and glasses from another), we extract the needed attribute from each reference image using its corresponding mask. These attribute regions are then normalized via cropping and spatially concatenated to form a single composite reference image of size $512 \times 512$.

\section{Further Analysis and Ablation Study}
\label{sec:suppl_analysis}

\subsection{Discussions on the generative model used in our Swapping-based Corruption.}
\label{sec:suppl_gen_model}
Our swapping-based corruption strategy consists of two steps:
1. Using a generative model to synthesize a new image from a real image, thereby corrupting the target attribute.
2. Applying a post-synthesis filter to discard synchronized pairs that exhibit poor preservation of non-target attributes.

\subsubsection{(a) Discussion on Qwen-Image-Edit.}
For the generative model, we typically utilize a general-purpose image editing model, Qwen-Image-Edit.

\begin{figure}[htbp]
  \begin{center}
    \includegraphics[width=  0.95  \textwidth]{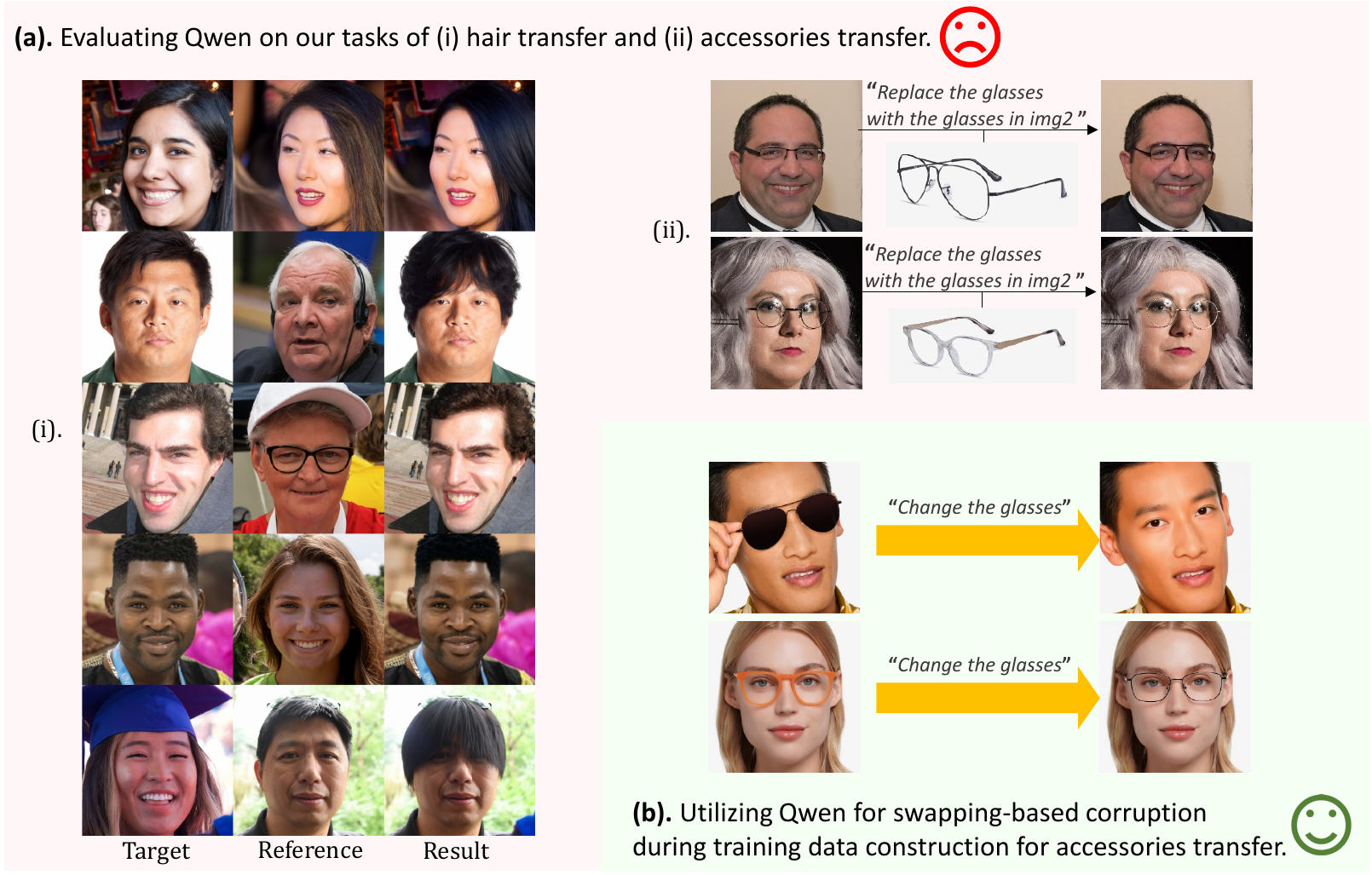}
  \end{center}
  \caption{
Analysis of the general image editing model Qwen-Image-Edit~\cite{qwen_image_edit}.
\textbf{(a)} Under the multiple-image mode (reference-based editing), Qwen-Image-Edit exhibits many unstable behavior like under-follow the reference, over-transfer and damage the target, or hallucinate in the edited region. (More examples are shown in Section~\ref{sec:suppl_vis}: Fig.~\ref{fig:suppl-hair}-~\ref{fig:suppl-face}). 
The consistently poor quantitative results shown in Tab.~1 further confirm its \textbf{unsuitability for the transfer task}.
\textbf{(b)} However, under the single-image mode (instruction-based editing), it can still generate plausible and diverse attribute variations, which is \textbf{sufficient for swapping-based corruption}.}
\label{fig:qwen}
\end{figure}

\textbf{Limitations in reference-based transfer.} \quad
As illustrated in Fig.~\ref{fig:qwen}-(a) and Section~\ref{sec:suppl_vis} (Fig.~\ref{fig:suppl-hair}-~\ref{fig:suppl-face}), Qwen-Image-Edit exhibits unstable, high-variance behavior across samples under the multiple-image mode (i.e., our reference-based transfer setting).
Specifically, it may under-follow the reference (a-i), over-transfer and damage the target (a-ii), or even hallucinate in the edited region (a-iii).
This high variability, together with its consistently low quantitative results in Tab.~\ref{tab:comp_qwen}, makes general-purpose multi-image editors unsuitable for our task.

\begin{table}[t]
    \raggedright
    \caption{Comparison with Qwen-Image-Edit (general-purpose image editing model).
    For each task, we report the most representative metric, that is identity similarity, hair clip distance, pose distance, and head clip distance.
    For all metrics, $\uparrow$ indicates higher is better and $\downarrow$ indicates lower is better.
    Best results are in \textbf{bold}.
}
    \vspace{-0.54em}
    \label{tab:comp_qwen}
    \center
    \resizebox{0.8 \linewidth}{!}{%
    \begin{tabular}{lcccc}
    \toprule
    \textbf{Method} & Face transfer$\uparrow$ & Hair transfer$\downarrow$ & Reenact$\downarrow$ & Head transfer$\downarrow$ \\
    \midrule
    Qwen-Image-Edit (arXiv 25) & 0.442 & 0.486 & 17.13 & 0.582 \\
    \textbf{Ours} & \textbf{0.637} & \textbf{0.421} & \textbf{7.09} & \textbf{0.460} \\
    \bottomrule
    \end{tabular}%
    }
\end{table}

For Tab.~\ref{tab:comp_qwen},
we use the following prompts (image 1: target, image 2: reference): \textit{(1). Face transfer:} ``Replace the face in image 1 with the face of image 2. Ensure 100\% fidelity to Image 2's facial identity. Only change face region. Strictly preserve the expression, head pose, hair, background, lighting, and skin tone of image 1.''\qquad \textit{(2). Hair transfer:} ``Replace the hair (style, color, and texture) in image 1 with the hair from image 2. Ensure 100\% fidelity to Image 2's hair. Only change hair region. Strictly preserve the face, head pose, and background of image 1.''\qquad \textit{(3). Motion transfer:} ``Use the head pose and facial expression of the subject in image 2 to drive the identity of the subject in image 1. Ensure 100\% fidelity to Image 2's pose and expression. Only alter head pose and facial expression in image 1. Preserve the face identity, hair, and background of image 1.''\qquad \textit{(4). Head transfer:} ``Replace the entire head (including face and hair) of the subject in image 1 with the head of the subject in image 2. Ensure 100\% fidelity to Image 2's head. Only change head region. Preserve the body and background of image 1.''

\textbf{Sufficiency for attribute corruption.} \quad
However, for our data corruption purposes, reference transfer is not required. As shown in Fig.~\ref{fig:qwen}-(b), the model under the single-image mode (instruction-based image editing) is capable of generating plausible and diverse attribute variations. 

\subsubsection{(b) Discussion on StableHair and self-evolution.}
For specific tasks such as hair and head transfer, where spatially-dynamic attributes primarily involve hair, we employ the task-specific model StableHair~\cite{stable_hair_aaai25} for efficiency.
As demonstrated in the visual comparisons for hair transfer (Fig.~5-(c) in the main paper and Fig.~\ref{fig:suppl-hair}), although StableHair displays suboptimal preservation of face lighting and background, our model, trained on data synchronized by StableHair, consistently achieves better preservation fidelity.

This observation highlights a key advantage of our framework: the data construction strategy effectively transforms imperfect generative models into stronger ones. By relaxing the requirements for the data generator (specifically demanding mainly diverse attribute variations rather than precise transfer) and filtering out low-quality outputs, we enable the final model to surpass the capabilities of the generator used to create its training data. 
This suggests the possibility of 'self-evolution': by bootstrapping from a suboptimal generative model to synchronize data, we can obtain an improved model, which can then be used to generate even better training data, leading to further model enhancements.

We perform a further experiment on the hair transfer task: we first obtain an initial model (the model presented in our main experiments) trained on data synthesized by the suboptimal generative model StableHair. We then leverage this initial model to construct higher-quality training data, which exhibits better background consistency between the ground truth and target images compared to the original StableHair data. Consequently, this evolved model achieves even greater preservation performance as shown in Tab.~\ref{tab:suppl_self_evolve}.

\begin{table}[t]
\centering
\caption{Quantitative results of the self-evolution experiment on hair transfer task. 
By leveraging our initial model to construct higher-quality training data, we can obtain an evolved model with improved performance.
}
\label{tab:suppl_self_evolve}
\resizebox{0.6\textwidth}{!}{%
\begin{tabular}{l|ccc}
\toprule
\textbf{Method} & CLIP dist.$\downarrow$ & ID sim$\uparrow$ & non-hair SSIM$\uparrow$ \\
\midrule
Stable-Hair (AAAI 25) & 0.468 & 0.855 & 0.87 \\
\underline{Our initial model} & \underline{0.421} & \underline{0.887} & \underline{0.91} \\
\textbf{Our evolved model} & \textbf{0.419} & \textbf{0.888} & \textbf{0.94} \\
\bottomrule
\end{tabular}%
}
\end{table}

\subsection{Quantitative ablation of Result-based Adaptive LoRA Rank Allocation.}
\label{sec:suppl_abla_lora}

\begin{table}[t]
\centering
\caption{We compare our Result-based Adaptive LoRA rank against the conventional gradient-based and the uniform rank. 
For each task, we report the most representative metric, that is identity similarity, hair clip distance, pose distance, and head clip distance.
}
\label{tab:rank_allocation_abla}
\resizebox{0.8\textwidth}{!}{%
\begin{tabular}{l|cccc}
\toprule
LoRA Rank Allocation Method & Face transfer$\uparrow$ & Hair transfer$\downarrow$ & Reenact$\downarrow$ & Head transfer$\downarrow$ \\
\midrule
Uniform & 0.636 & 0.419 & 10.13 & 0.460 \\
Gradient-based & 0.634 & 0.433 & 7.35 & 0.464 \\
Result-based (\textbf{Ours}) & 0.637 & 0.421 & 7.09 & 0.460 \\
\bottomrule
\end{tabular}%
}
\vspace{-1em}
\end{table}

In Tab.~\ref{tab:rank_allocation_abla}, we compare our Result-based Adaptive LoRA Rank Allocation with uniform rank across tasks per layer, and gradient-based adaptive rank, while maintaining a consistent total parameter count.
While the other three tasks maintain high performance, motion transfer suffers a \textbf{significant decline}. 
Without enough task-specific capacity for the motion transfer task, the model is heavily influenced by the other three tasks, which are all about transfering some visual elements from reference to target. As a result, it now struggles to drive the target with reference motion during motion transfer.
This finding is corroborated by the specific rank distribution: 
taking the first MoE-FFN layer in the first middle block of the main UNet as an example, the assigned ranks for the face transfer, hair transfer, motion transfer and head transfer are 49, 37, 112 and 48 respectively. That is to say, our strategy assigns the highest rank (i.e., 112) to the motion transfer task, whereas the ranks for the other three tasks remain below 50. This indicates that motion transfer weights diverge significantly from other tasks.
Our result-based approach effectively addresses this cross-task difference by automatically identifying and assigning higher ranks to the most conflict-prone tasks.

\subsection{Examples of discarded and retained pairs in data construction}
\label{sec:suppl_data_filter}
\begin{figure}[htbp]
  \begin{center}
    \includegraphics[width=  1  \textwidth]{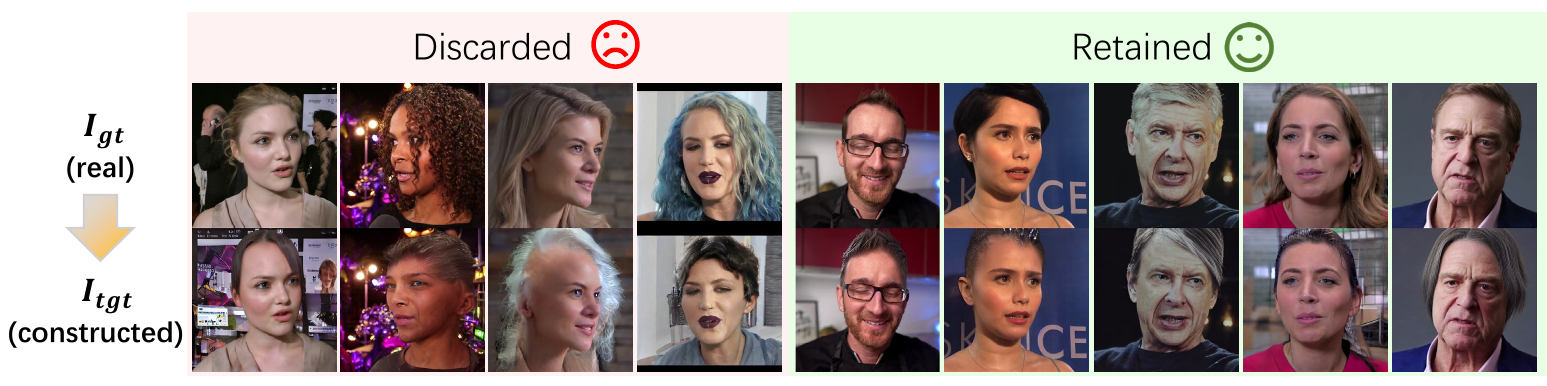}
  \end{center}
  \caption{
Examples of discarded and retained image pairs after filtering for hair transfer. (a) Discarded pairs due to low background SSIM, low face identity similarity, large pose L2 distance, and large expression L2 distance, respectively; (b) Retained pairs after filtering.  
}
\label{fig:filtered_examples}
\end{figure}

Fig.~\ref{fig:filtered_examples} shows examples of retained and discarded image pairs after the filtering step for the hair transfer task. Specifically, pairs are discarded when the ground-truth image $I_{\text{gt}}$ and the constructed target image $I_{\text{tgt}}$ exhibit low background SSIM, low face identity similarity, large pose L2 distance, or large expression L2 distance.
We use percentile-based thresholds rather than absolute values. For each metric, we sort samples from best to worst (for similarity style metrics, higher is better; for distance style metrics, lower is better) and keep the top $Fraction_{\text{metric}}$ fraction. Any sample that fails any metric is discarded (union of rejects), so the final yield is the proportion remaining after all filters are applied. For example, for hair transfer, we use
$Fraction_{\text{face\_sim}}=0.8$ (face identity similarity),
$Fraction_{\text{pose\_dist}}=0.8$ (pose L2 distance),
$Fraction_{\text{exp\_dist}}=0.8$ (expression L2 distance),
and $Fraction_{\text{bg\_sim}}=0.5$ (background SSIM).
These values reflect that pose is the most quality-sensitive factor, while expression and background tolerate looser cutoffs. We log the retained ratio for each dataset split in our filtering script and will release the code with these exact settings for reproducibility.

\subsection{Discussion on StyleGAN-based Methods}
\label{sec:suppl_stylegan}

Prior to the proliferation of diffusion models, StyleGAN-based latent manipulation~\cite{styleflow, stylespace, barbershop} was the dominant paradigm for high-resolution portrait editing. Methods like StyleFlow~\cite{styleflow}, StyleSpace~\cite{stylespace}, and Barbershop~\cite{barbershop} achieve impressive results by exploring disentangled latent spaces for attribute editing and image compositing. 

\begin{figure*}[htbp] 
    \centering
    \includegraphics[width=0.7\linewidth]{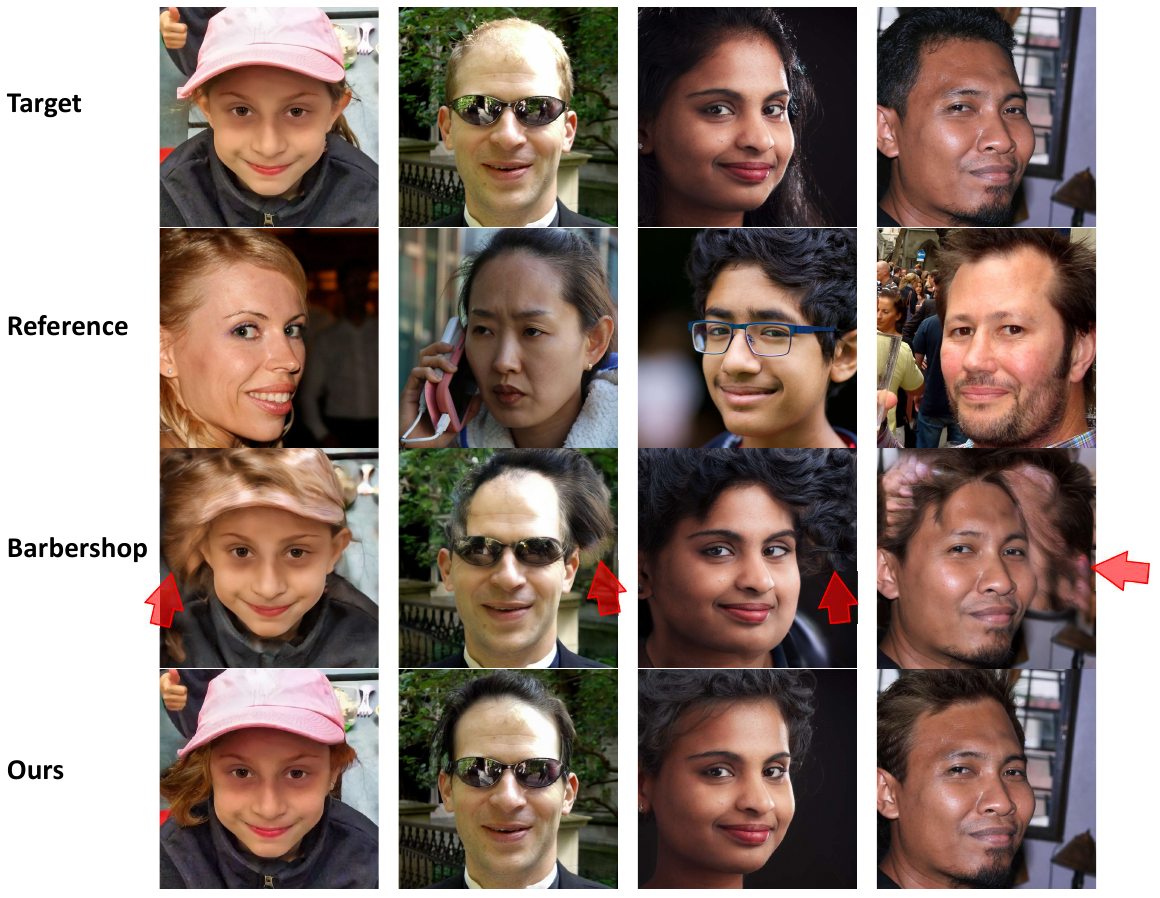} 
    \caption{
\textbf{Qualitative comparison against the StyleGAN-based method, Barbershop~\cite{barbershop}.} 
As indicated by the red arrows, Barbershop encounters obvious visual artifacts and distortions when there is a pose discrepancy between the reference and target images. Moreover, due to its heavy reliance on latent optimization, it requires $\ge 4$ minutes to infer a single image pair, rendering it highly unscalable and unsuitable for our unified, efficient multi-task generation task.
    }
    \label{fig:bbs}
\end{figure*}

Despite these significant achievements, they suffer from severe limitations in real-world scenarios compared to more recent feed-forward methods. While these methods exhibit excellent performance when both the reference and target images share strictly aligned, frontal poses, they struggle significantly with pose discrepancies. In practical applications, reference and target images are rarely strictly aligned and typically exhibit different head poses. Under such unconstrained conditions, methods like Barbershop (see Fig.~\ref{fig:bbs}) fail to geometrically adapt the spatial structure; they either rigidly copy and paste the reference attributes, or generate glaring artifacts and distorted boundaries, as shown in Fig.~\ref{fig:bbs}. 
Furthermore, to achieve an acceptable blending outcome, these methods rely on computationally heavy test-time latent optimization, which takes $\ge 4$ minutes per image pair. Such extreme inefficiency makes them highly unscalable and impractical compared to the single feed-forward pass of our proposed method.

\section{More Quantitative Comparison}
\label{sec:suppl_quant}
\subsection{More metrics for head transfer}
\label{sec:suppl_metrics}
In the main paper, we only show some key metrics for head transfer to save space. Here we show more metrics in 
Tab.~\ref{tab:suppl_head_replacement_extended}.




\subsection{Efficiency comparison}
\label{sec:suppl_efficiency}
To complement the quality metrics, we also report computational efficiency in Tab.~\ref{tab:suppl_efficiency}, including peak CUDA memory and inference time per image.
Our method achieves competitive efficiency among unified methods while maintaining strong transfer performance across multiple tasks.

\begin{table}[t]
    \raggedright
    \caption{Quantitative comparison with more metrics for head transfer. 
For all metrics, $\uparrow$ indicates higher is better and $\downarrow$ indicates lower is better. 
Best result from Unified Methods are marked \textcolor{red}{Red}.
Overall best results are highlighted in \textbf{bold}, while overall second-best results are \underline{underlined}.
}
    \vspace{-0.5em}
    \label{tab:suppl_head_replacement_extended}
    \resizebox{\linewidth}{!}{%
    \begin{tabular}{lcccccc}
    \toprule
    \textbf{Method} & CLIP dist.$\downarrow$ & ID sim$\uparrow$ & pose dist.$\downarrow$ & expr. dist.$\downarrow$ & CLIP dist. (hair)$\downarrow$ & FID$\downarrow$ \\
    \midrule
    \multicolumn{7}{l}{\textbf{Task-Specific Methods}} \\
    GHOST2.0 (arXiv 25) & \underline{0.618} & 0.461 & 5.78 & 1.867 & \underline{0.497} & 31.38 \\
    \midrule
    \multicolumn{7}{l}{\textbf{Multi-Task Methods (training separately for different tasks)}} \\
    REFace (WACV 25) & 0.639 & \underline{0.540} & \underline{5.43} & \underline{1.776} & 0.549 & \underline{8.33} \\
    \midrule
    \multicolumn{7}{l}{\textbf{Unified Methods (train once for all tasks)}} \\
    Ours & \textbf{\textcolor{red}{0.460}} & \textbf{\textcolor{red}{0.545}} & \textbf{\textcolor{red}{5.28}} & \textbf{\textcolor{red}{1.772}} & \textbf{\textcolor{red}{0.399}} & \textbf{\textcolor{red}{8.28}} \\
    \bottomrule
    \end{tabular}%
    }
\end{table}

\begin{table}[t]
    \raggedright
    \caption{Efficiency comparison: Runtime and Peak CUDA memory. 
    For all metrics, $\downarrow$ indicates lower is better. 
    For multi-task methods, we report the maximum runtime and peak CUDA memory across all tasks.}
    \vspace{-0.5em}
    \label{tab:suppl_efficiency}
    \center
    \resizebox{0.9\linewidth}{!}{%
    \begin{tabular}{lcc}
    \toprule
    \textbf{Method} & Peak CUDA memory (GB)$\downarrow$ & Infer time (second/img)$\downarrow$ \\
    \midrule
    \multicolumn{3}{l}{\textbf{Task-Specific Methods}} \\
    CanonSwap (ICCV 25) & 11.7 & $<1$ \\
    Stable-Hair (AAAI 25) & 19.3 & 16 \\
    HairFusion (AAAI 25) & 10.9 & 22 \\
    HunyuanPortrait (CVPR 25) & 18.4 & 103 \\
    GHOST2.0 (arXiv 25) & 8.0 & $<1$ \\
    \midrule
    \multicolumn{3}{l}{\textbf{Multi-Task Methods (training separately for different tasks)}} \\
    REFace (WACV 25) & 11.8 & 12 \\
    \midrule
    \multicolumn{3}{l}{\textbf{Unified Methods (train once for all tasks)}} \\
    Face-Adapter (ECCV 24) & 6.8 & 3 \\
    RigFace (arXiv 25) & 21.9 & 18 \\
    Ours & 14.5 & 15 \\
    \bottomrule
    \end{tabular}%
    }
\end{table}

\subsection{Comparisons under complex scenes: extreme poses, exaggerated expressions, and occlusions}
\label{sec:suppl_complex_scenes}
We show quantitative comparisons under complex scenes in Tab.\ref{tab:complex_all} and visual comparisons in Fig.\ref{fig:complex_t0}, Fig.\ref{fig:complex_t1}, Fig.\ref{fig:complex_t2}, and Fig.\ref{fig:complex_t3}.
To construct the visual comparison set, we manually select images with occlusions from the FFHQ dataset~\cite{ffhq}. Because exaggerated expressions and extreme poses are rare in FFHQ, we source representative samples from the AffectNet dataset~\cite{affectnet} and the ExtremePose-Face-HQ (EFHQ) dataset~\cite{dao2024efhq}, respectively.

For the quantitative analysis, we also evaluate performance across three distinct scenarios: exaggerated expressions using AffectNet, extreme poses using EFHQ, and occlusions using the Real World Occluded Faces (ROF) dataset~\cite{rof_dataset}. For each scenario, we sample 333 images and pair each with a standard image from the FFHQ test set, yielding 333 test pairs per setting. 
When computing metrics, as some approaches fail to process certain images due to brittle preprocessing steps (e.g., failed face detection), we omit these failure cases from the affected methods' evaluations to ensure fair metric computation.

As shown in Tab.~\ref{tab:complex_all}, Fig.~\ref{fig:complex_t0}, Fig.~\ref{fig:complex_t1}, Fig.~\ref{fig:complex_t2} and Fig.~\ref{fig:complex_t3}, all methods naturally suffer performance degradation under these highly challenging conditions. Nevertheless, our method still achieves the best overall performance.

\begin{table*}[t]
    \centering
    \caption{Comparisons on the most competitive methods for each task under complex scenes.}
    \label{tab:complex_all}
    \resizebox{\textwidth}{!}{%
    \begin{tabular}{l|l|ccc|ccc|ccc}
    \toprule
    \textbf{Task} & \textbf{Method} & \multicolumn{3}{c|}{\textbf{Exaggerated Expressions}} & \multicolumn{3}{c|}{\textbf{Extreme
Pose}} & \multicolumn{3}{c}{\textbf{Occlusions}} \\
    \cmidrule(lr){3-5} \cmidrule(lr){6-8} \cmidrule(lr){9-11}
    & & \multicolumn{3}{c|}{ID sim$\uparrow$ \quad pose dist.$\downarrow$ \quad expr. dist.$\downarrow$}
    & \multicolumn{3}{c|}{ID sim$\uparrow$ \quad pose dist.$\downarrow$ \quad expr. dist.$\downarrow$}
    & \multicolumn{3}{c}{ID sim$\uparrow$ \quad pose dist.$\downarrow$ \quad expr. dist.$\downarrow$} \\
    \midrule
    \multirow{3}{*}{Face Transfer}
    & REFace & \underline{0.509} & 5.10 & 1.28         & \underline{0.293} & 10.09 & 1.10      & \underline{0.311} & 5.93 & 1.07 \\
    & CanonSwap & 0.446 & \textbf{2.27} & \textbf{0.91}      & 0.141 & \textbf{3.73} & \underline{0.75}      & 0.152 & \textbf{2.53} & \textbf{0.76} \\
    & Ours & \textbf{0.518} & \underline{3.54} & \underline{1.04}                     & \textbf{0.302} & \underline{4.25} & \textbf{0.73}      & \textbf{0.316} & \underline{4.80} & \underline{1.06} \\
    \midrule
    & & CLIP dist.$\downarrow$ & ID sim$\uparrow$ & non-hair SSIM$\uparrow$
    & CLIP dist.$\downarrow$ & ID sim$\uparrow$ & non-hair SSIM$\uparrow$
    & CLIP dist.$\downarrow$ & ID sim$\uparrow$ & non-hair SSIM$\uparrow$ \\
    \cmidrule(lr){3-5} \cmidrule(lr){6-8} \cmidrule(lr){9-11}
    \multirow{2}{*}{Hair Transfer}
    & Stable-Hair & \underline{0.460} & \underline{0.869} & \underline{0.871} & \textbf{0.477} & \underline{0.831} & \underline{0.880} & \underline{0.472} & \underline{0.826} & \underline{0.848} \\
    & Ours & \textbf{0.436} & \textbf{0.904} & \textbf{0.931} & \textbf{0.477} & \textbf{0.899} & \textbf{0.941} & \textbf{0.452} & \textbf{0.883} & \textbf{0.930} \\
    \midrule
    & & ID sim$\uparrow$ & pose dist.$\downarrow$ & expr. dist.$\downarrow$
    & ID sim$\uparrow$ & pose dist.$\downarrow$ & expr. dist.$\downarrow$
    & ID sim$\uparrow$ & pose dist.$\downarrow$ & expr. dist.$\downarrow$ \\
    \cmidrule(lr){3-5} \cmidrule(lr){6-8} \cmidrule(lr){9-11}
    \multirow{3}{*}{Motion Transfer}
    & Face-Adapter              & \underline{0.492} & \textbf{7.75} & \underline{2.31}    & \underline{0.337} & \textbf{11.93} & \underline{2.60}    & \underline{0.472} & \underline{10.61} & \underline{2.46} \\
    & HunyuanPortrait           & 0.357 & 8.39 & \underline{2.42}    & 0.237 & 13.48 & 2.65    & \underline{0.484} & 11.84 & 2.56 \\
    & Ours                      & \textbf{0.497} & \underline{8.16} & \textbf{2.27}    & \textbf{0.338} & \underline{12.63} & \textbf{2.21}    & \textbf{0.509} & \textbf{10.15} & \textbf{2.32} \\
    \midrule
    & & CLIP dist.$\downarrow$ & ID sim$\uparrow$ & pose dist.$\downarrow$
    & CLIP dist.$\downarrow$ & ID sim$\uparrow$ & pose dist.$\downarrow$
    & CLIP dist.$\downarrow$ & ID sim$\uparrow$ & pose dist.$\downarrow$ \\
    \cmidrule(lr){3-5} \cmidrule(lr){6-8} \cmidrule(lr){9-11}
    \multirow{3}{*}{Head Transfer}
    & GHOST2.0            & \underline{0.495} & 0.369 &  6.07  & \underline{0.525} & 0.257 & \underline{8.32}   & \underline{0.512} & 0.332 & \underline{8.15} \\
    & REFace              & 0.552 & \underline{0.568} & \underline{5.87}   & 0.559 & \underline{0.465} & 14.56   & 0.555 & \underline{0.458} & 11.54 \\
    & Ours                & \textbf{0.416} & \textbf{0.573} & \textbf{5.20}   & \textbf{0.479} & \textbf{0.484} & \textbf{8.02}   & \textbf{0.424} & \textbf{0.490} & \textbf{7.43} \\
    \bottomrule
    \end{tabular}%
    }
\end{table*}

\section{More Visual Comparison}
\label{sec:suppl_vis}

\begin{figure*}[htbp]
  \begin{center}
    \includegraphics[width=1.0\textwidth]{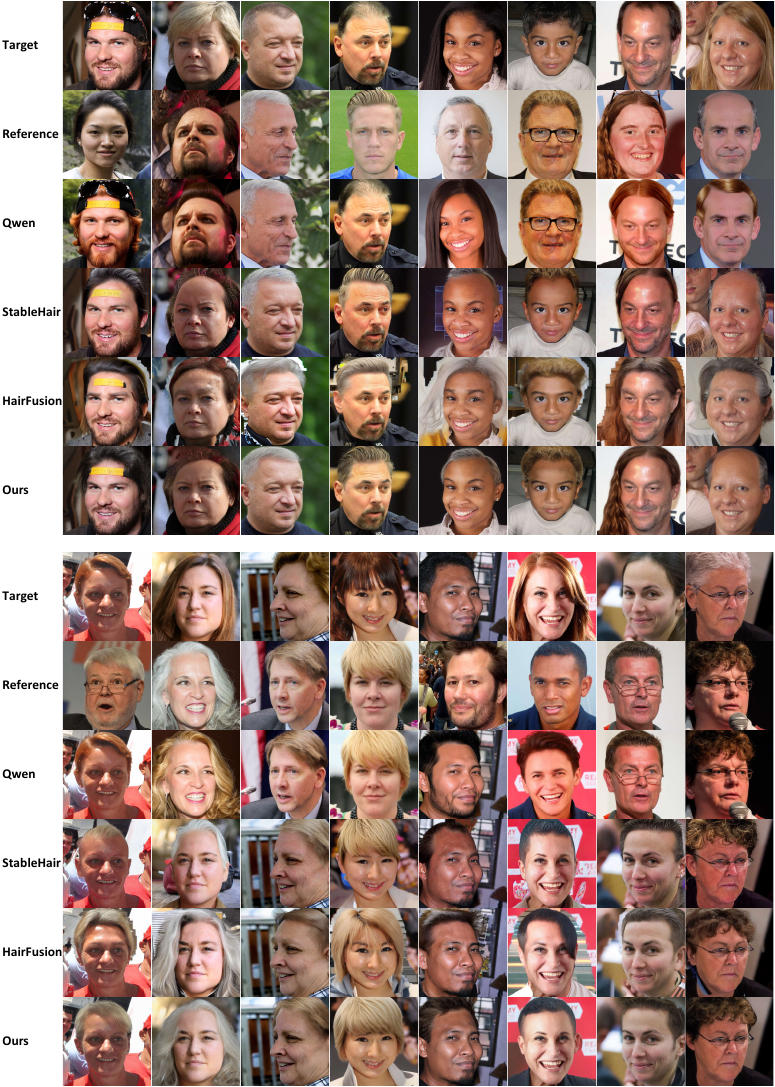}
  \end{center}
  \caption{Hair transfer visual comparison.}\label{fig:suppl-hair}
\end{figure*}

\begin{figure*}[htbp]
  \begin{center}
    \includegraphics[width=1.0\textwidth]{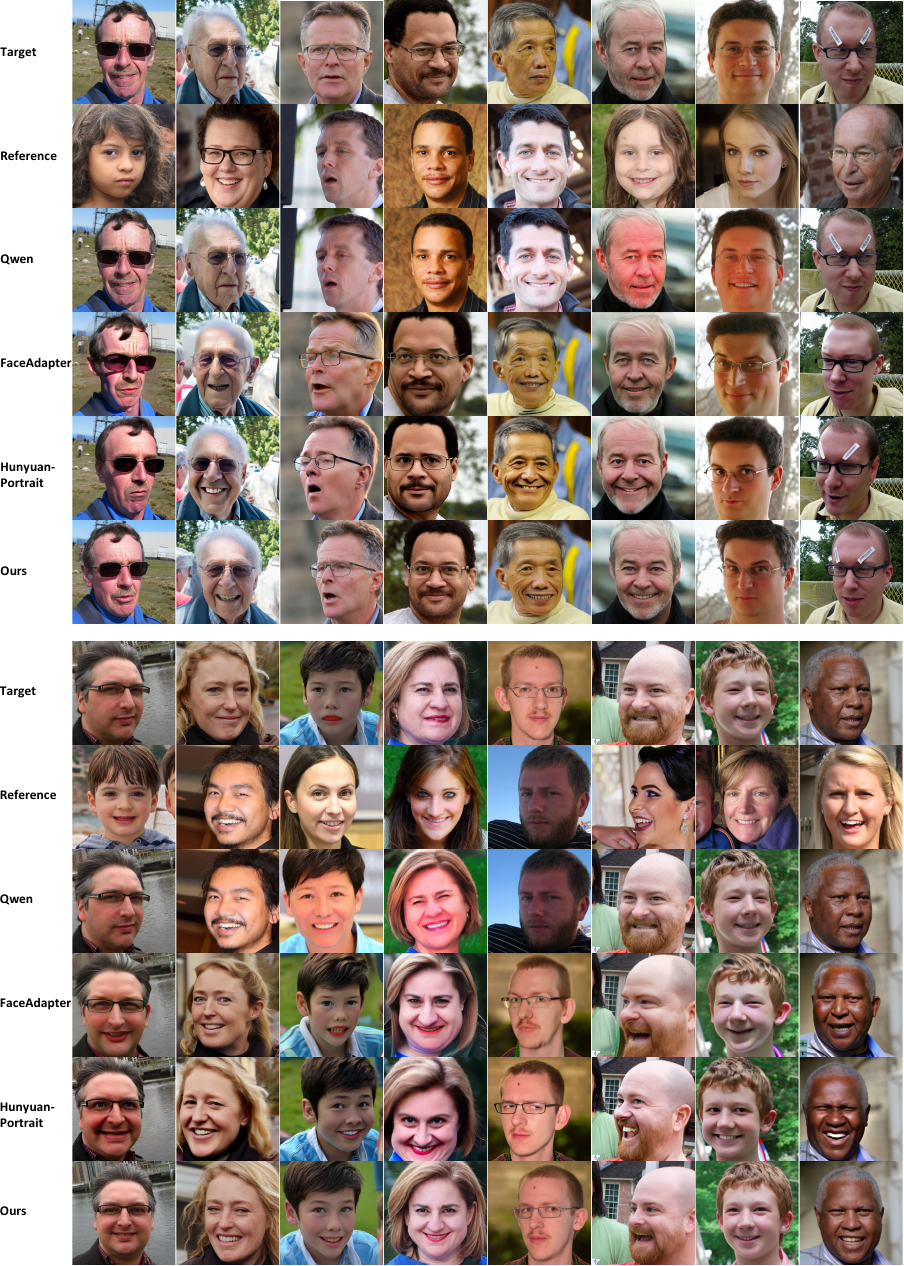}
  \end{center}
  \caption{Motion transfer visual comparison.
HunyuanPortrait tends to change the face skin, or lighting. 
The faceAdapter shows suboptimal face fidelity. 
}\label{fig:suppl-reen}
\end{figure*}

\begin{figure*}[htbp]
  \begin{center}
    \includegraphics[width=1.0\textwidth]{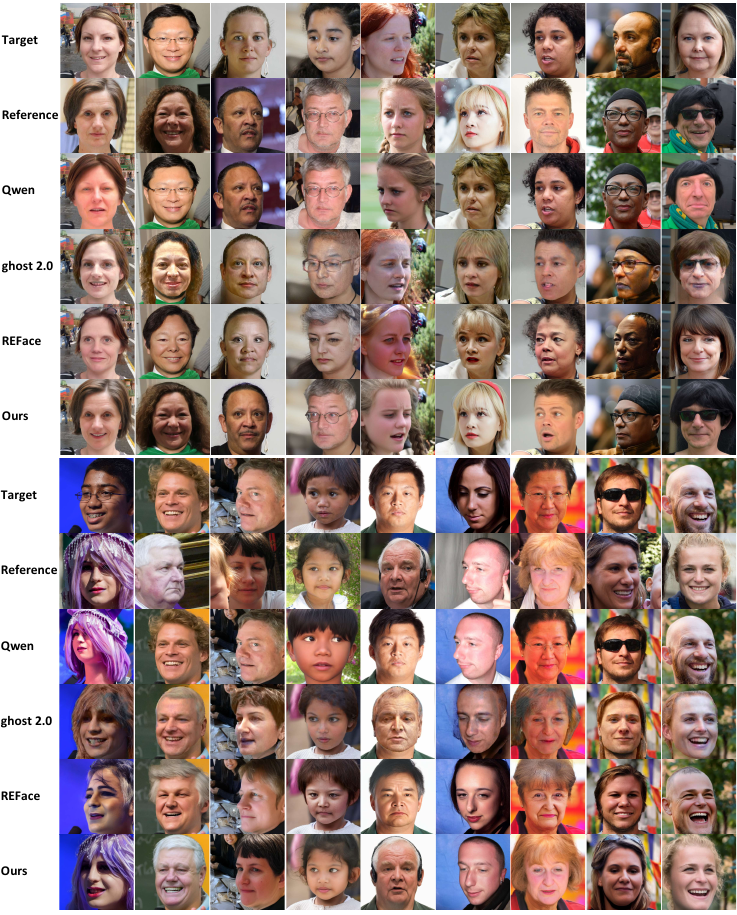}
  \end{center}
  \caption{Head transfer visual comparison.}\label{fig:suppl-head}
\end{figure*}

\begin{figure*}[htbp]
  \begin{center}
    \includegraphics[width=0.99\textwidth]{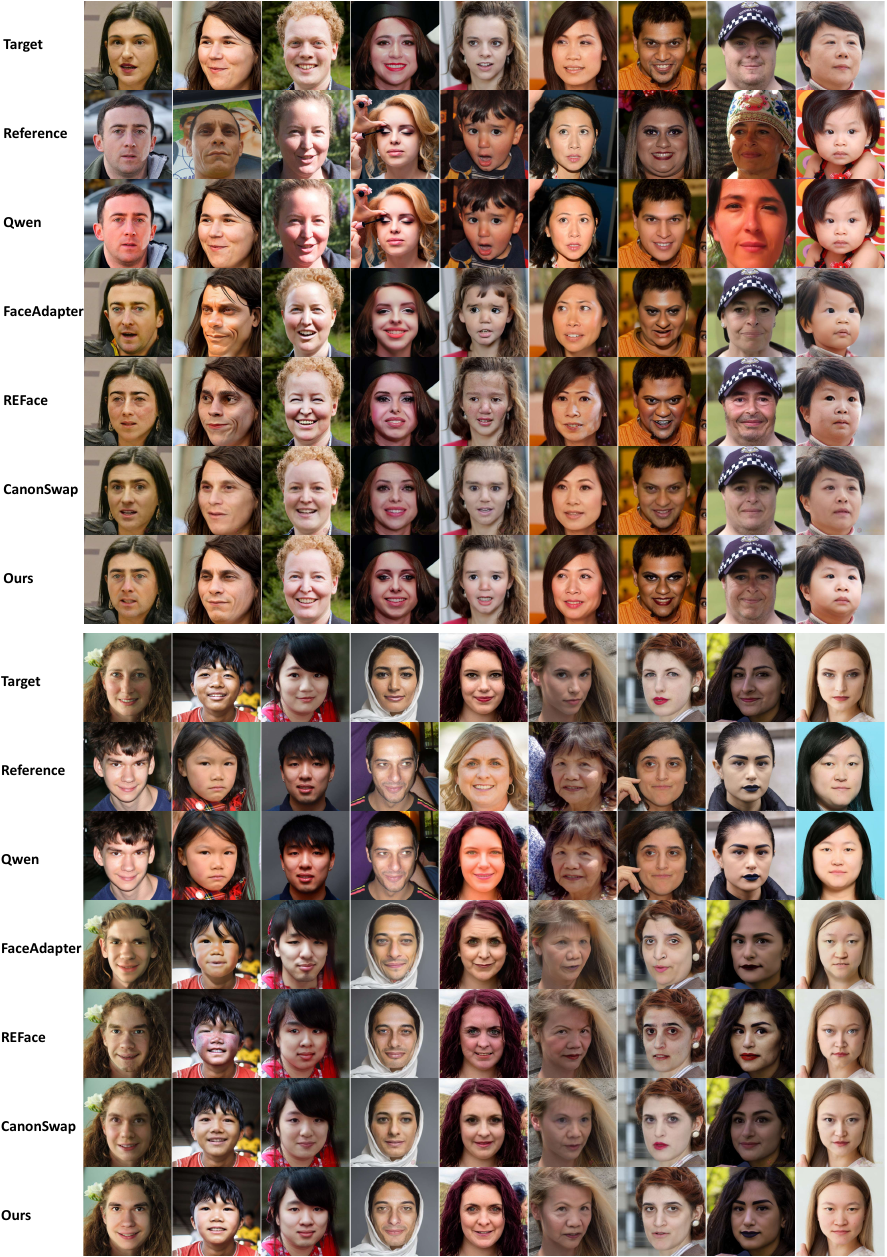}
  \end{center}
  \vspace{-1.2em}
  \caption{Face transfer visual comparison.}\label{fig:suppl-face}
\end{figure*}

\begin{figure*}[htbp]
  \begin{center}
    \includegraphics[width=0.99\textwidth]{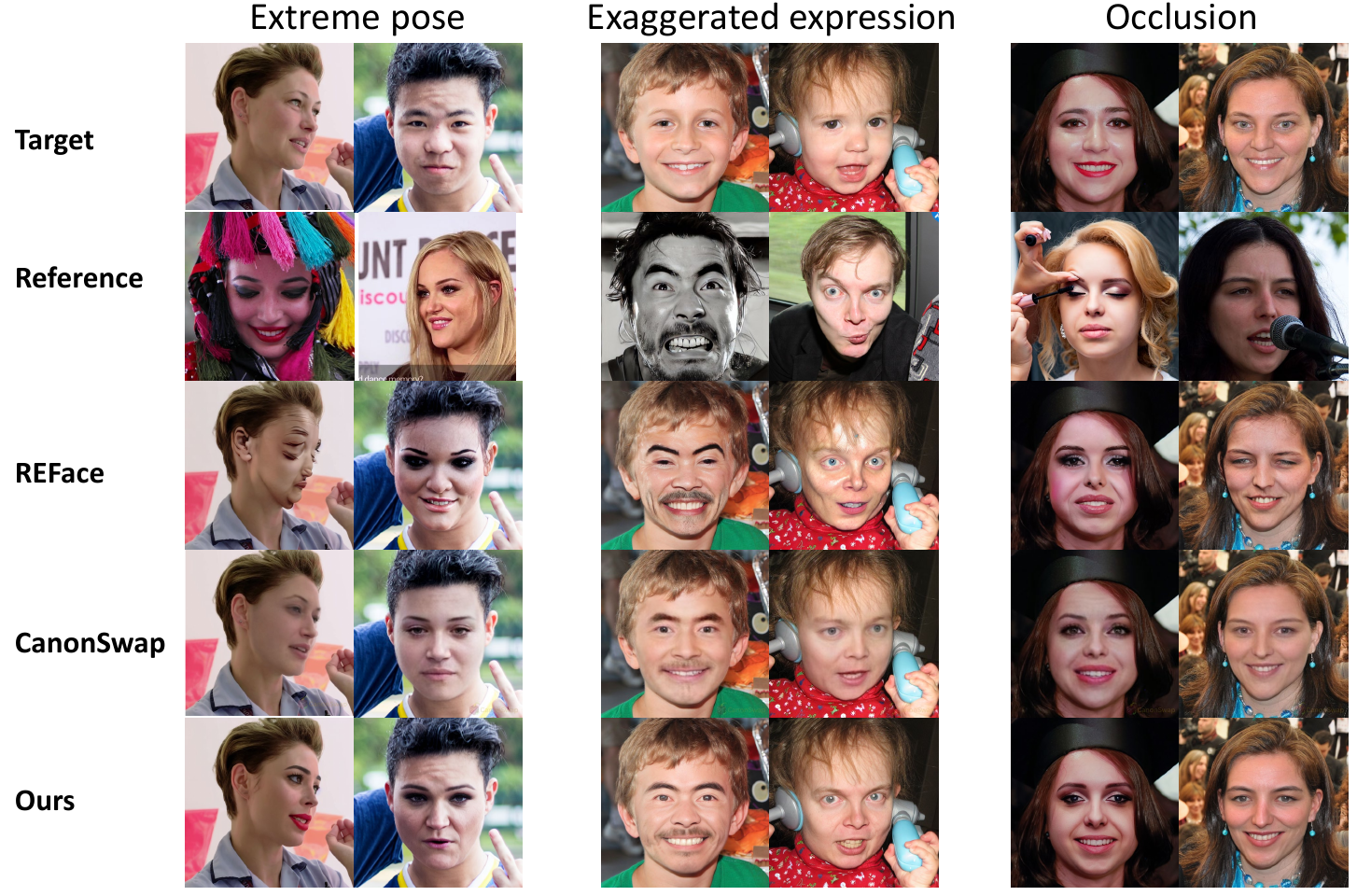}
  \end{center}
  \vspace{-1.2em}
  \caption{Visual comparison of \textbf{face transfer} under complex scenes.}\label{fig:complex_t0}
\end{figure*}

\begin{figure*}[htbp]
  \begin{center}
    \includegraphics[width=0.99\textwidth]{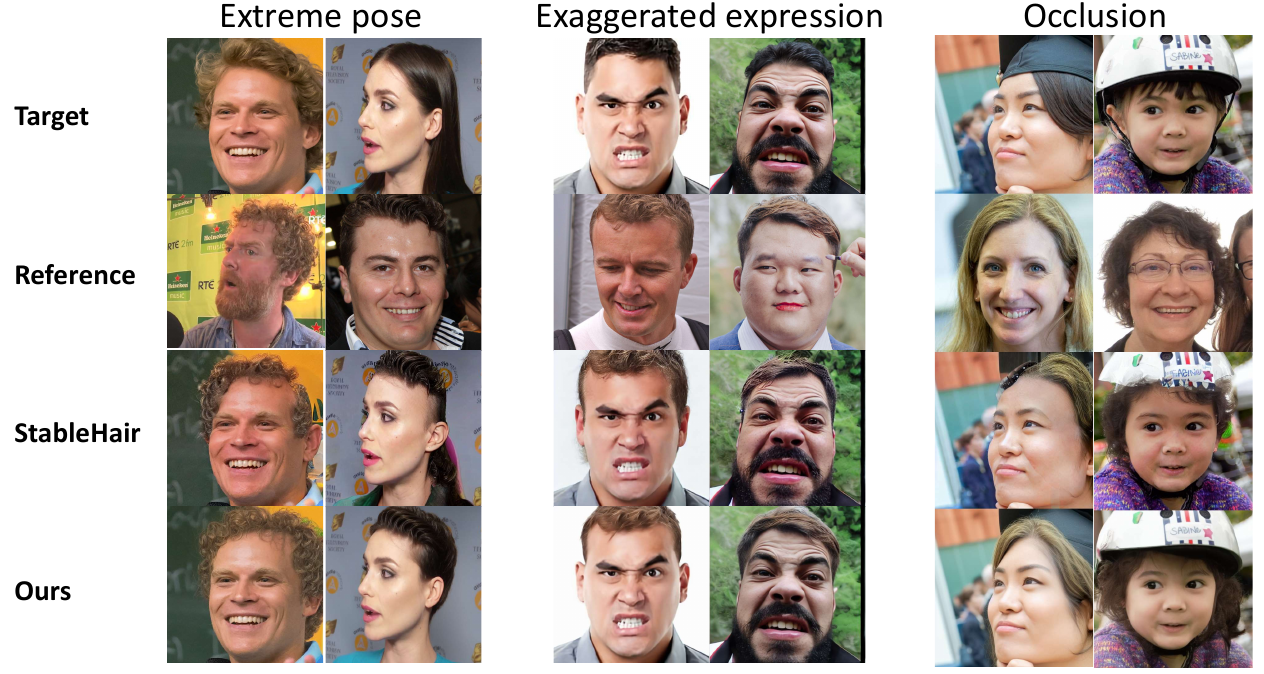}
  \end{center}
  \vspace{-1.2em}
  \caption{Visual comparison of \textbf{hair transfer} under complex scenes.}\label{fig:complex_t1}
\end{figure*}

\begin{figure*}[htbp]
  \begin{center}
    \includegraphics[width=0.99\textwidth]{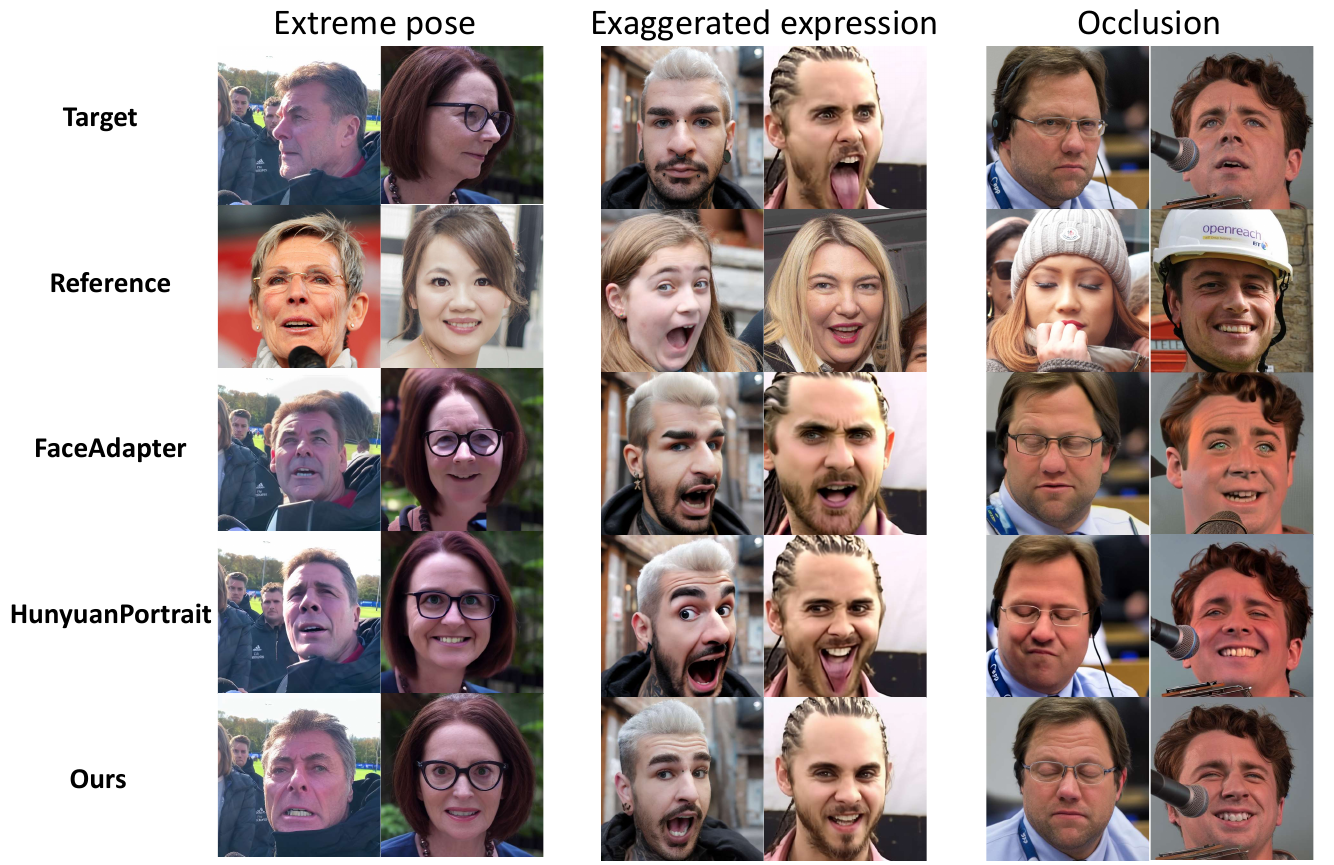}
  \end{center}
  \vspace{-1.2em}
  \caption{Visual comparison of \textbf{motion transfer (face reenactment)} under complex scenes.}\label{fig:complex_t2}
\end{figure*}

\begin{figure*}[htbp]
  \begin{center}
    \includegraphics[width=0.99\textwidth]{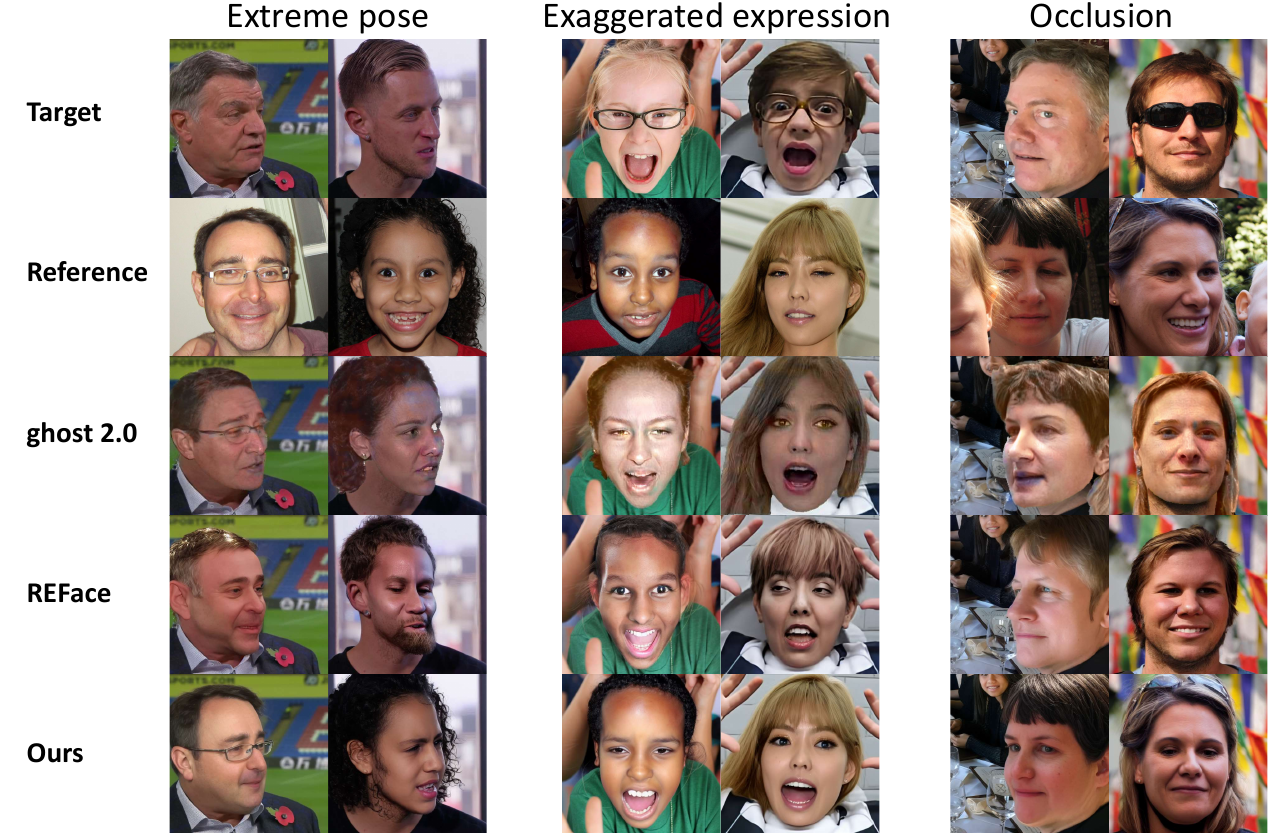}
  \end{center}
  \vspace{-1.2em}
  \caption{Visual comparison of \textbf{head transfer} under complex scenes.}\label{fig:complex_t3}
\end{figure*}

We provide more visual results in Fig.~\ref{fig:suppl-hair}, Fig.~\ref{fig:suppl-reen}, Fig.~\ref{fig:suppl-head}, and Fig.~\ref{fig:suppl-face}.

\section{Limitations}
\label{sec:suppl_limit}
Despite the improvements achieved by \textbf{UniBioTransfer}, some limitations remain:
\begin{itemize}
    \item \textbf{Performance on Asian faces.} Our model sometimes exhibits suboptimal performance on certain ethnic groups, particularly Asian faces. This may be mostly attributed to the data distribution of the training datasets, which is also faced by many works in the field.
    \item Due to computational resource constraints, we only conducted experiments using Stable Diffusion v1.5 as the backbone. Theoretically, our entire framework, including the unified data construction, BioMoE and two-stage training strategy, can be applied to other backbones like FLUX. We leave these explorations for future work.
\end{itemize}

\clearpage
\bibliographystyle{splncs04}
\bibliography{main}

\end{document}